%% file: main.tex
\documentclass[letterpaper,twocolumn,10pt, dvipsnames]{article}
\usepackage{usenix2019_v3}

\usepackage{tikz}
\usepackage{amsmath}
\usepackage{algorithm}
\usepackage{algorithmic}
\usepackage{url}
\usepackage{fancyvrb}
\usepackage{listings}
\usepackage{xcolor}
\usepackage{multirow}
\usepackage{booktabs}
\usepackage{caption}
\usepackage{amsmath}
\usepackage{subcaption}

\newcommand{\ra}[1]{\renewcommand{\arraystretch}{#1}} 
\newcommand*{\Comb}[2]{{}^{#1}C_{#2}}%

\usepackage{colortbl}
\definecolor{lightgray}{gray}{0.9}

\definecolor{codegreen}{rgb}{0,0.6,0}
\definecolor{codegray}{rgb}{0.5,0.5,0.5}
\definecolor{codepurple}{rgb}{0.58,0,0.82}
\definecolor{backcolour}{rgb}{0.95,0.95,0.92}

\usepackage{filecontents}

\lstdefinestyle{mystyle}{
    backgroundcolor=\color{backcolour},   
    commentstyle=\color{codegreen},
    keywordstyle=\color{magenta},
    numberstyle=\tiny\color{codegray},
    stringstyle=\color{codepurple},
    basicstyle=\ttfamily\footnotesize,
    breakatwhitespace=false,         
    breaklines=true,                 
    captionpos=b,                    
    keepspaces=true,                 
    numbers=left,                    
    numbersep=5pt,                  
    showspaces=false,                
    showstringspaces=false,
    showtabs=false,                  
    tabsize=1,
    frame=tb, 
    frameround=tttt,
    linewidth=1\linewidth, 
    columns=fullflexible, 
}

\AtBeginDocument{%
  \providecommand\BibTeX{{%
    \normalfont B\kern-0.5em{\scshape i\kern-0.25em b}\kern-0.8em\TeX}}}

\input{header}
\input{terminologies}

\iftrue
\newcommand{\sm}[1]{\todo[inline,color=brown!40]{Subrata: #1}}
\newcommand{\sa}[1]{\todo[inline,color=green!40]{Shubham: #1}}
\newcommand{\sarc}[1]{\todo[inline,color=blue!40]{Sarthak: #1}}

\else
\newcommand{\sm}[1]{}
\newcommand{\sa}[1]{}
\newcommand{\sarc}[1]{}

\fi

\lstdefinestyle{mystyle}{
    backgroundcolor=\color{gray!20},
    basicstyle=\footnotesize\ttfamily,
    frame=single,
    framerule=0.5pt, 
    framesep=2pt, 
    rulecolor=\color{gray!50}, 
}

\newcommand*\circled[1]{\tikz[baseline=(char.base)]{
            \node[shape=circle,fill,inner sep=0.3pt] (char) {
             \footnotesize \bf{ 
            \textcolor{white}{#1}
             }
            } ;}}

\newcommand*\circledw[1]{\tikz[baseline=(char.base)]{
            \node[shape=circle, draw=black, inner sep=0.3pt] (char) {
            \footnotesize \bf{ 
            \textcolor{black}{#1}
             }
            } ;}}

\DeclareMathOperator*{\argmin}{arg\,min}

\begin{document}

\title{Approximate Caching for Efficiently Serving Diffusion Models
}

\author{
{Shubham Agarwal} \\ Adobe Research
\and
{Subrata Mitra\thanks{Corresponding author.}} \\
Adobe Research
\and
{Sarthak Chakraborty\thanks{Work done at Adobe Research.}}\\
UIUC
\and
{Srikrishna Karanam}\\
Adobe Research
\and
{Koyel Mukherjee}\\
Adobe Research
\and
{Shiv Kumar Saini}\\
Adobe Research
\and
}


\maketitle

\input{tex/abstract.tex}
\input{tex/introduction.tex}

\input{tex/background.tex}

\input{tex/analysis.tex}
\input{tex/overview.tex}

\input{tex/design.tex}

\input{tex/implementation.tex}

\input{tex/evaluation.tex}

\input{tex/related_work.tex}

\input{tex/conclusion.tex}

{\footnotesize
\bibliographystyle{plain}
\bibliography{main}
}

\end{document}

%% file: header.tex
\usepackage{kantlipsum,setspace,titlesec}
\usepackage{todonotes}

\usepackage{tikz}
\usepackage{amsmath}
\usepackage{amsthm}
\usepackage{amssymb}
\usepackage{xspace}
\usepackage{xurl}
\usepackage{filecontents}
\usepackage{cleveref}
\usepackage{comment}
\usepackage{graphicx}
\usepackage[font={it,small},labelsep=colon, belowskip=-1.0pt, aboveskip=0.3pt]{caption}
\usepackage[shortlabels]{enumitem}
\setlist[enumerate]{nosep}
\setlist{nolistsep,leftmargin=4.0mm}
\usepackage{mathtools}
\usepackage[hang,flushmargin]{footmisc} 
\usepackage{xcolor}
\usepackage{algorithm}
\usepackage{algorithmic}
\usepackage{kantlipsum,setspace,titlesec}
\usepackage{subcaption}
\usepackage{flushend}

\usepackage{xpatch}
\newcommand{\ie}{{i.e.,}\xspace}
\newcommand{\eg}{{e.g.,}\xspace}
\newcommand{\etal}{\textit{et al.}\xspace}

%% file: terminologies.tex
\newcommand{\sys}[0]{\textsc{Nirvana}\xspace}
\newcommand{\sysomp}[0]{\textsc{Nirvana}-w/o\textsc{MP}\xspace}
\newcommand{\policy}[0]{\texttt{LCBFU}\xspace}

\newcommand{\gpt}[0]{\textsc{GPT-Cache}\xspace}
\newcommand{\pinecone}[0]{\textsc{Pinecone}\xspace}
\newcommand{\clip}[0]{\textsc{CRS}\xspace}

\newcommand{\vanilla}[0]{\textsc{Vanilla}\xspace}

\newcommand{\DiffusionDB}[0]{\textsc{DiffusionDB}\xspace}

\newcommand{\smallmodel}[0]{\textsc{SmallModel}\xspace}

\newcommand{\X}{{\sc System-X}\xspace}

\newcommand{\dms}[0]{\text{diffusion-models}\xspace}

\newcommand{\embeddinggenerator}[0]{\texttt{embedding-generator}\xspace}

\newcommand{\matchpredictor}[0]{\texttt{match-predictor}\xspace}

\newcommand{\VDB}[0]{\texttt{VDB}\xspace}
\newcommand{\vdb}[0]{\texttt{VDB}\xspace}

\newcommand{\cacheselector}[0]{\texttt{cache-selector}\xspace}

\newcommand{\cachemaintainer}[0]{\texttt{cache-maintainer}\xspace}

\newcommand{\efs}[0]{\texttt{EFS}\xspace}

\newcommand{\dm}[0]{\texttt{diffusion-model}\xspace}

%% file: tex/abstract.tex
\begin{abstract}
Text-to-image generation using diffusion models has seen explosive popularity owing to their ability in producing high quality images adhering to text prompts.
However, diffusion-models go through a large number of iterative denoising steps, and are resource-intensive, requiring expensive GPUs and incurring considerable latency.
In this paper, we introduce a novel \textit{approximate-caching} technique that can reduce such iterative denoising steps by reusing intermediate noise states created during a prior image generation.
Based on this idea, we present an end-to-end text-to-image generation system, \sys, that uses approximate-caching with a novel cache management policy to provide 21\% GPU compute savings, 19.8\% end-to-end latency reduction and 19\% dollar savings on two real 
production workloads. We further present an extensive characterization of real production text-to-image prompts from the perspective of caching, popularity and reuse of intermediate states in a large production environment.  
\end{abstract}

%% file: tex/introduction.tex
\section{Introduction}
\begin{figure}[t]
    \centering
    \begin{subfigure}[b]{0.48\linewidth}
        \includegraphics[width=1\columnwidth]{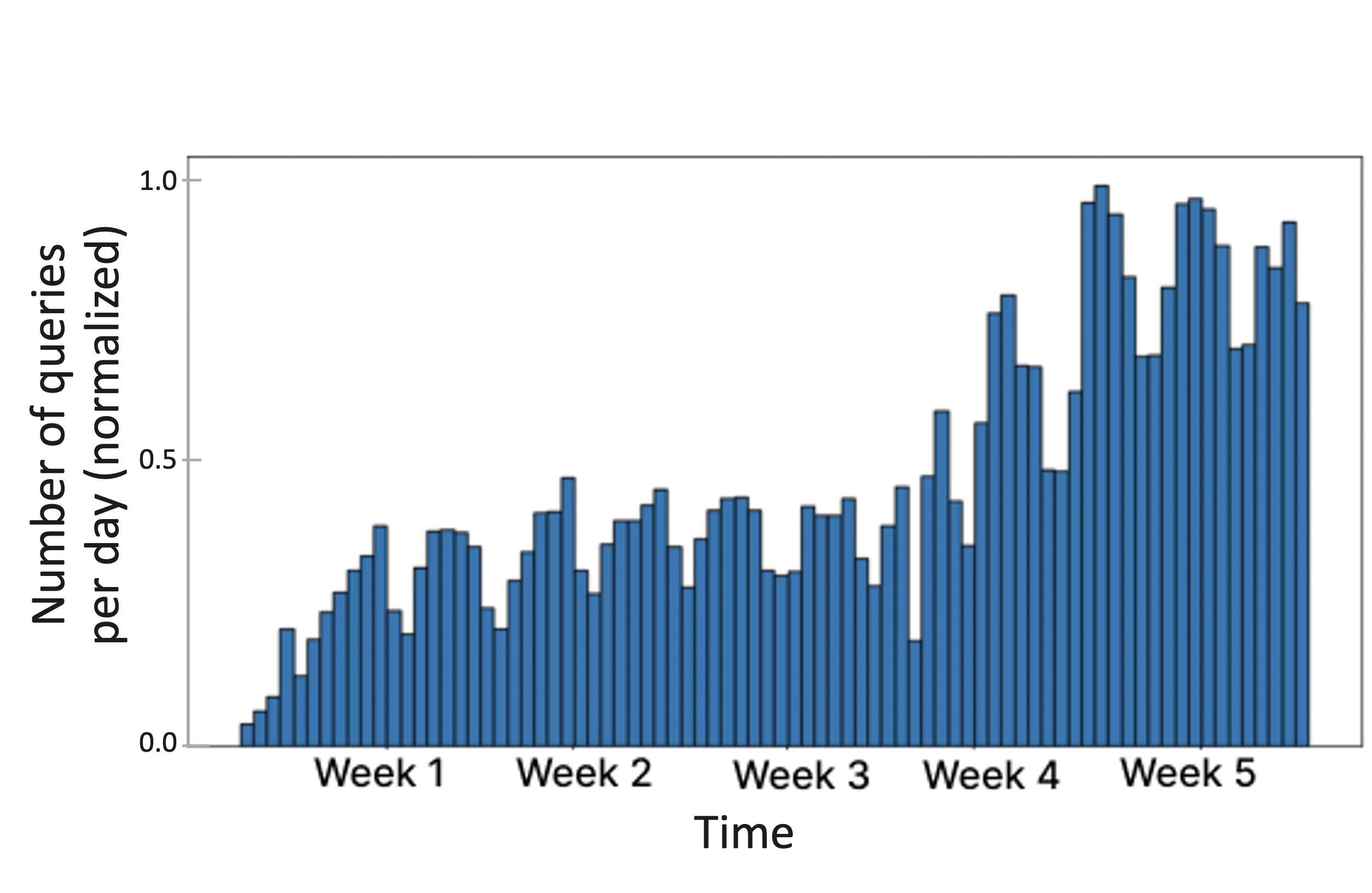}
        \caption{}
        \label{fig:growth-diffusion}
    \end{subfigure}
    \begin{subfigure}[b]{0.48\linewidth}
        \includegraphics[width=1\columnwidth]{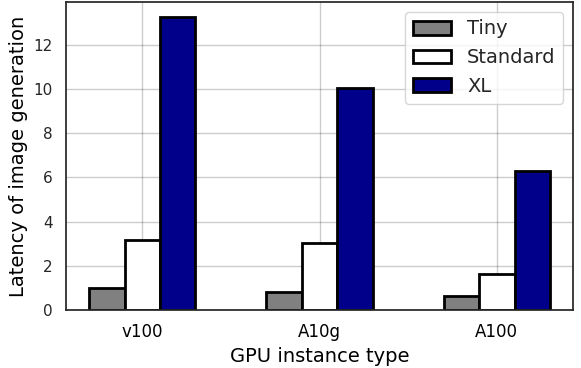}
        \caption{}
        \label{fig:aws_cost}
    \end{subfigure}
    \caption{(a) Shows normalized growth in workload over 5 weeks in DiffusionDB. (b) Latency for \texttt{tiny}, \texttt{standard} and \texttt{XL} stable-diffusion models from Hugging Face~\cite{huggingface} repository on three different GPU architectures in AWS. The image quality produced by the larger \texttt{XL} model is far superior to the smaller two models but comes with a substantially large latency overhead which also implies higher costs.}
    \label{fig:enter-label}
\end{figure}

Text-to-image generation has drastically matured over the years~\cite{zhang2023text, croitoru2023diffusion} and has now become a widely popular feature offered by various companies~\cite{intel_lab, adobe_express_usage}, being integrated in various new creative workflows~\cite{firefly_news}.
The popularity of such a text-to-image models has become massive. 
Adobe recently reported~\cite{firefly_news} that over 2 billion images were created using its Firefly~\cite{firefly} text-to-image service. 
Similar popularity has also been reported for Dall-E-2 from OpenAI~\cite{dall-e}. 
Figure~\ref{fig:growth-diffusion} shows the staggering growth over time in the numbers of prompts submitted to a portal running stable-diffusion-based text-to-image model as captured by the DiffusionDB dataset~\cite{wang2022diffusiondb}.

\begin{figure}[t]
    \centering
    \begin{subfigure}[b]{\linewidth}
         \centering
         \includegraphics[width=0.9\textwidth]{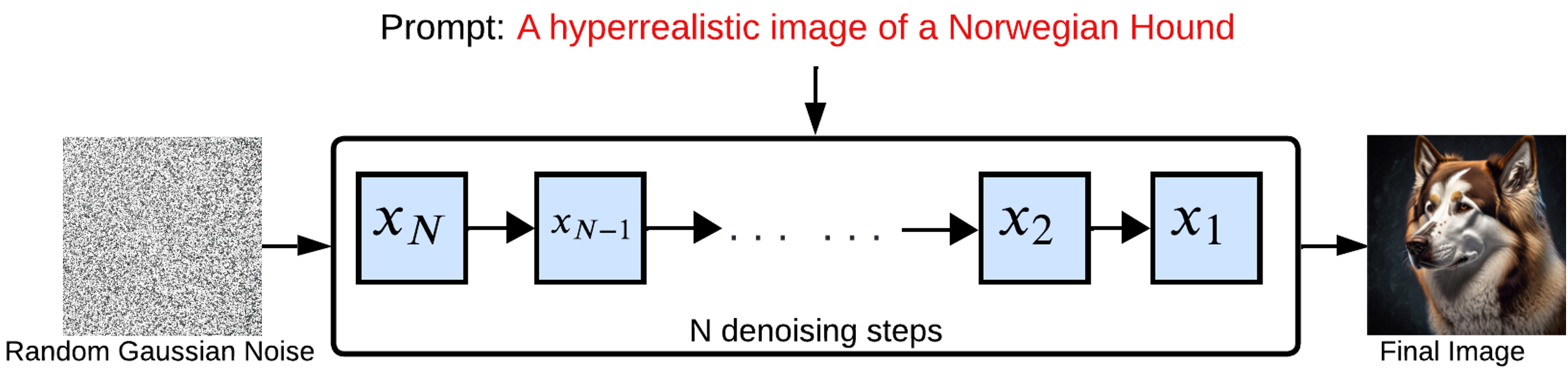}
         \caption{}
         \label{fig:wo_cache}
    \end{subfigure}
    \begin{subfigure}[b]{\linewidth}
         \centering
         \includegraphics[width=1.0\textwidth]{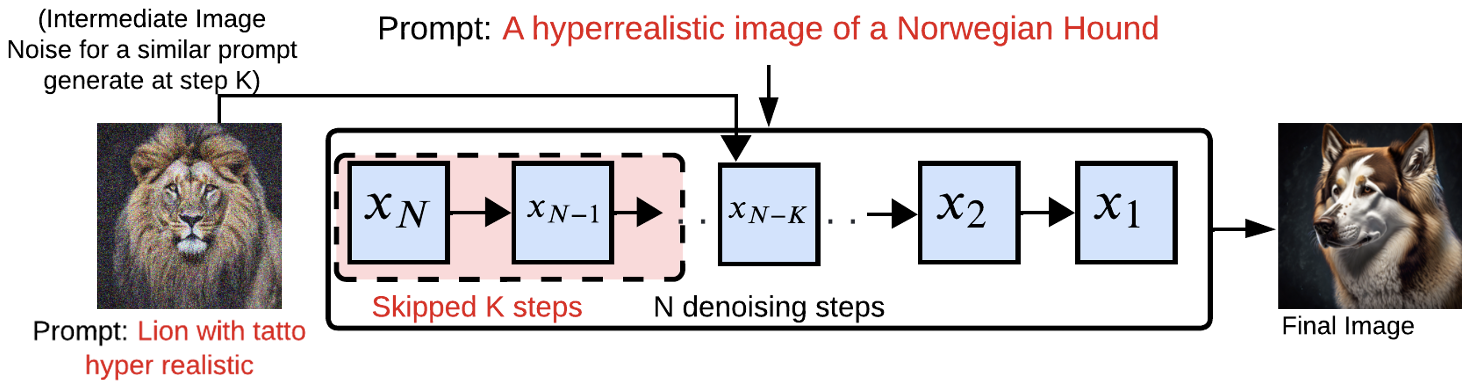}
         \caption{}
         \label{fig:with_cache}
    \end{subfigure}
    \caption{(a) Shows how vanilla \dms work. (b) Shows how \dms with approximate-caching works where first $K$ denoising steps are skipped after an intermediate-noise belonging to a different prompt present in the cache is retrieved and reused.}
    \label{fig:diffusion_example}
\end{figure}

\textbf{State-of-the-art in text-to-image: }In Text-to-image generation, given a text \textit{prompt}, describing certain desirable characteristics of an image, a deep neural network model generates an image capturing the descriptions provided in the prompt. 
While researchers have been attempting to design viable and consistent text-to-image models for quite some time, using VAEs~\cite{kingma2013auto}, GANs~\cite{NIPS2014_5ca3e9b1} and other techniques~\cite{dosovitskiy2020image}, the most popular text-to-image production systems of today~\cite{firefly, midjourney} are based on \textit{Diffusion-Models}~\cite{dhariwal2021diffusion, ho2020denoising}. 
The wide-spread adaptation of \dms can be attributed to their capability to generate superior quality images and to condition the generation more accurately according to the input prompt.

\textbf{Problems with \dms: } Image generation using \dms is a computationally resource intensive task and suffers from relatively long latency. During the inference phase in \dms (that is generating an image from text), essentially a Gaussian noise is iteratively denoised, using the input prompt as a condition, to produce the output image in a Markov process~\cite{gagniuc2017markov}. 
The traditional \dms use as much 1000 such iterative steps~\cite{dhariwal2021diffusion} for this. 
Some subsequent optimizations on this process~\cite{liu2022pseudo, song2020denoising} enables such denoising to be done with approximately 50-100 iterative steps~\cite{Complete85:online}. 

However, even with 50 diffusion steps, image generation with \dms is a high resource intensive and slow task that prohibits interactive experience for the users and results in huge computational cost on expensive GPUs.

In Figure~\ref{fig:aws_cost} we show the latency (in seconds) for different GPU architectures (\ie A100, A10g and V100 from NVIDIA) that are available on Amazon Web Services (AWS) and for few \dms of different sizes (\ie \# of parameters): \texttt{Stable Diffusion XL}, \texttt{Stable Diffusion 2.1}, \texttt{Tiny Stable Diffusion} from Hugging Face repository~\cite{huggingface} with typical 50 diffusion steps for image generation during inference. 
It can be observed that while smaller models (\eg \texttt{Tiny} and \texttt{Standard}) can provide significantly low inference latency compared to a larger model(\eg \texttt{XL}), it usually comes with significant degradation in the quality of the generated image~\cite{podell2023sdxl}. 
Therefore, reducing a smaller model to reduce latency for better user experience might actually defeat the purpose. 
Latency can also be reduced by using a more powerful GPU for inference as can be observed in Figure~\ref{fig:aws_cost}, but pricing for cloud instances with powerful GPUs such as NVIDIA A100s are significantly high. 
For instance, in the US East region\cite{AWSProdu62:online} the V100 is priced at approximately \$3.06 per hour, the A10g at \$8.144 per hour, and the A100 at \$32.77 per hour making a A100 instance $4X$ costlier than A10g and more than $10X$ costlier than a V100. 
Therefor while inference latency can be reduced with more powerful GPUs, inference cost per image also significantly increases.

In this paper, we introduce an efficient text-to-image generation system called \sys that uses a novel technique called \textit{approximate  caching} to significantly reduce computational cost and latency by intelligently reusing intermediate states created during image generation for prior prompts.

Figure~\ref{fig:diffusion_example} illustrates the key idea behind \sys. For an input prompt shown in \textit{red} in Figure~\ref{fig:wo_cache} we show how the vanilla or standard \dms work through multiple iterative denoising steps. Here the process starts from a Gaussian noise at state $x_N$ and then performs $N$ denoising steps with the input prompt as the condition to finally produce a coherent image in state $x_0$. 
Figure~\ref{fig:with_cache} shows how in approximate-caching first $K$ steps are skipped and directly a suitable retrieved intermediate-noise from a different prompt is used. 
The value of $K$ depends on the similarity between the new input prompt and the prompt from which the noise was retrieved. Therefore, the amount of compute and latency savings can vary across prompts based on availability of similar prompts in the cache that stores the intermediate noises/states produced during the image generation from previously encountered prompts.
The phrase \textit{approximate} caching emphasizes the fact that in this system, we are not directly reusing the retrieved object from the cache, rather we are retrieving an intermediate state and further conditioning those to tailor the generated image according to the new prompt. 
Thus \sys is very different from a retrieval-based system such as \gpt~\cite{zillizte97:online}, \pinecone~\cite{MakingSt85:online} that proposes to directly retrieve an image from a cache based on the input prompt.
\begin{figure}[t]
  \includegraphics[width=1.0\columnwidth]{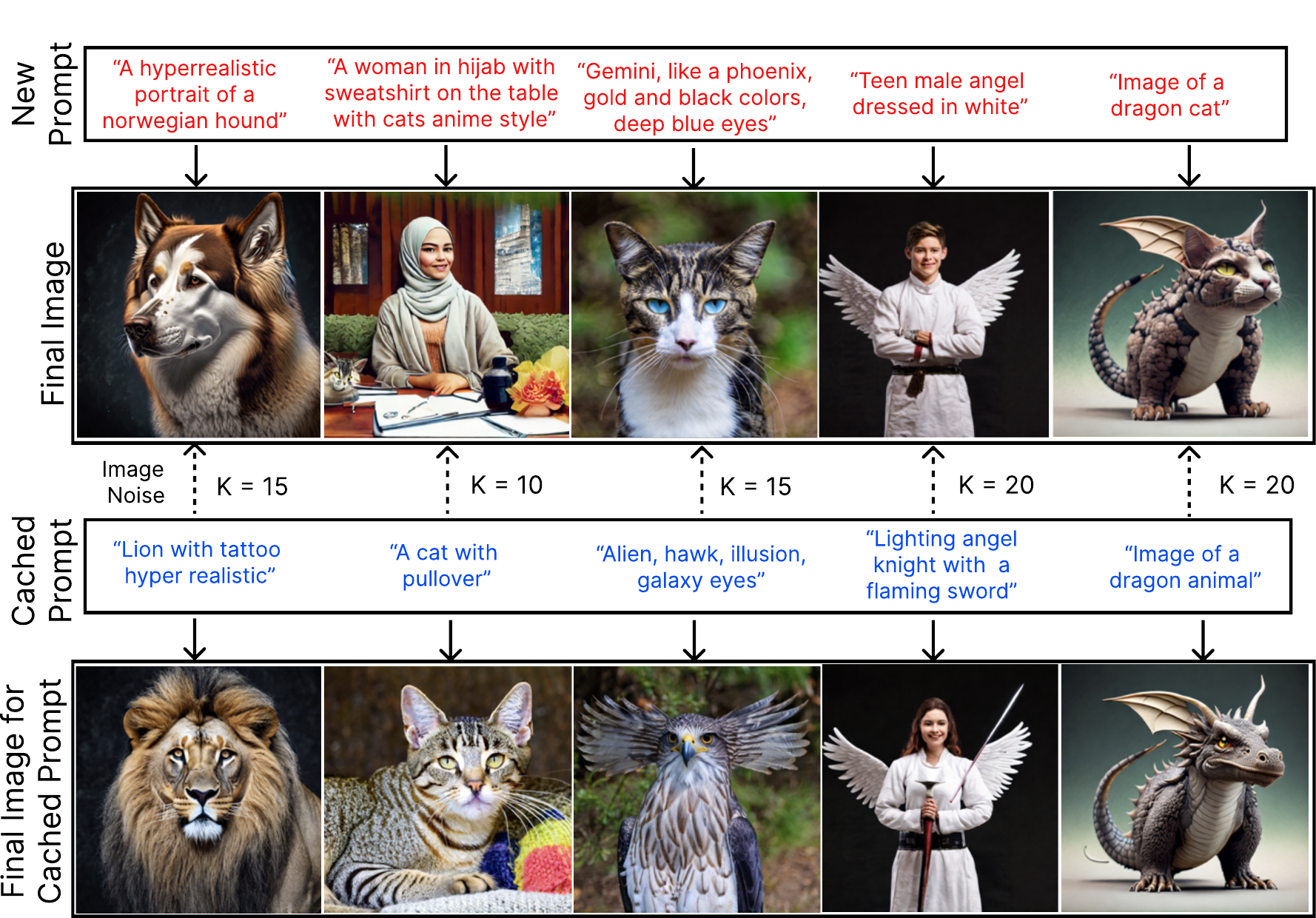}
  \caption{Images generated using \sys
  \label{fig:wow}
  }
\end{figure}

The heuristic design in \sys for selecting the value of $K$ for a particular input prompt effectively controls the \textit{hit-rate} vs. compute savings trade-off. Hit-rate here means how likely it is that an incoming prompt will find a \textit{similar enough} match in the cache and $K$ denotes the \# steps we skip at the beginning of the \dms while using the retrieved noise from the cache. 
The hit-rate, can be made significantly high if we are planning to skip very small number of steps at the beginning (\ie lower value for $K$).
The reason being, if we are skipping less sampling steps at the beginning noise from very dissimilar prompts can still be conditioned effectively to produce a coherent image. While for high $K$ the scope for modifying the latent space of the noise according to new prompt becomes limited, but when for the prompts it can be done, can provide huge compute savings. Careful design of \sys can navigate this complexity by calculating a suitable value for $K$ for each incoming prompt maintain high-rate as well optimize compute saving.

 Figure~\ref{fig:wow} shows some real examples using prompts from DiffusionDB~\cite{wang2022diffusiondb} to illustrate how \sys can transform a noise from a seemingly different prior prompt to a coherent and high-quality image while providing significant latency and compute reduction at the same time.

Furthermore, in \sys we design a novel approximate-cache management policy, called \textit{Least Computationally Beneficial and Frequently Used} (\policy),
that manages the storage noises in such a manner that for a given cache storage size, we optimize the space for the noises that is likely to give the best computational efficiency to \sys.

Overall, \sys can maintain generated image quality very close to vanilla diffusion-models (\ie without approximate-caching) while providing 21 \% savings in GPU computation, 19.8 \% reduction in latency and 19\% amortized savings on dollars spent on image generation.
Our study involving 60 users with 1000 images show that 79\% users liked \sys generated images which is far better than the best baseline \gpt~\cite{zillizte97:online} with only 31\% likes, and much closer to the quality of images generated by the expensive and slow \vanilla model, liked by 86\% users.

We summarize our contributions in this paper as follows:

\begin{enumerate}
    \item We introduce the novel idea of \textit{approximate caching} that provides significant computation saving in the production pipeline of diffusion models for text-to-image generation. 
    \item We propose an effective cache-management mechanism called \policy, that can optimize the reuse of computation states and computation savings.
    \item We present end-to-end design details and rational for \sys, which is our optimized text-to-image deployment system on cloud.
    \item We characterize real production prompts for text-to-image models from the perspective of reusability and caching. 
    \item We present extensive evaluation with two real and large production prompts from text-to-image models along with a human evaluation and several sensitivity studies.
\end{enumerate}

%% file: tex/background.tex
\section{Background}
\subsection{Diffusion Models}
\label{sec:background_analytical}
Diffusion model progressively denoises a random Gaussian noise to generate an image conditioned on text. The training procedure contains a forward diffusion process, which obscures an image by adding noise repeatedly in a Markov process until it saturates to a Gaussian noise. In the backward diffusion process, original image is recovered by removing noise repeatedly. Each such denoising step is called ``sampling" since the model generates a sample by removing noise, and the method used for sampling is called the \textit{sampler}. 

In the forward diffusion process, Gaussian noise gets progressively added to an initial image $x_0$ for $T$ steps to get $x_T$. With the Markov chain assumption, it is expressed as:

\begin{equation}
q(x_{1:T}|x_0) := \prod_{t=0}^{T} q(x_t|x_{t-1})
\end{equation}

\begin{equation}
q(x_t | x_{t-1}) := \mathcal{N}(x_t | \sqrt{1 - \beta_t} x_{t-1}, \beta_t I)
\end{equation}

where $q(x_t | x_{t-1})$ is the posterior probability, and $\beta_1, ..., \beta_T$ is the noise schedule (either learned or fixed) to regulate the noise level at each diffusion step. Similarly, the backward diffusion process can be written as: 

\begin{equation}
\label{bakward_diff}
p_\theta(x_{0:T}) := p(x_T)\prod_{t=0}^{T} p_\theta(x_{t-1}|x_{t})
\end{equation}

\begin{equation}
p_\theta(x_{t-1} | x_{t}) := \mathcal{N}(x_{t-1} |  \bar{\mu_t}, \bar{\beta_t} I)
\end{equation}

where $p_\theta(.)$ denotes the probability of observing $x_{t-1}$ given $x_t$. Here, $p(x_T) = \mathcal{N}(x_T | 0, I)$. Here, $\bar{\mu_t}$ and $\bar{\beta_t}$ is learned. The objective is to learn $p(\theta)$ that maximizes the likelihood of training data in the backward/reverse diffusion process. Recent optimizations approximate the backward diffusion process by skipping certain intermediate states at predetermined timesteps ~\cite{zheng2023fast, kong2021fast}, thus reducing inference steps from $T\approx 1000$ steps to $N\approx 50$ steps. This is achieved by learning a sampler that predicts how much noise will remain after $T/N$ step for every one diffusion step. Even with these optimizations, image generation still take 10 seconds on A10g and 6 seconds on
A100 GPUs for large models (Figure~\ref{fig:aws_cost}).

Diffusion model backbone is based on the U-Net architecture~\cite{ronneberger2022convolutional}. When a text prompt is given, the image generation process is conditioned through cross-attention within the model~\cite{rombach2022high}. Thus, it develops a text-to-image framework capable of generating visually coherent and contextually relevant images based on textual descriptions. 

In practice, each $x_i$ can either be the actual image, or its latent representation computed by an image encoder~\cite{rombach2022high}. Generally, later approach, termed latent diffusion model (LDM) is preferred since it captures the hidden characteristics of an image. We use the same in our work. To generate the final image from its latent space $x_0$ at the end of backward diffusion, $x_0$ is decoded with the inverse of the image encoder.

\subsection{Dynamics of Image Generation}
\label{sec:Iteration_analysis}

\begin{figure*}[t]
 \centering
 \begin{minipage}{0.33\textwidth}
   \includegraphics[width=1.0\textwidth]{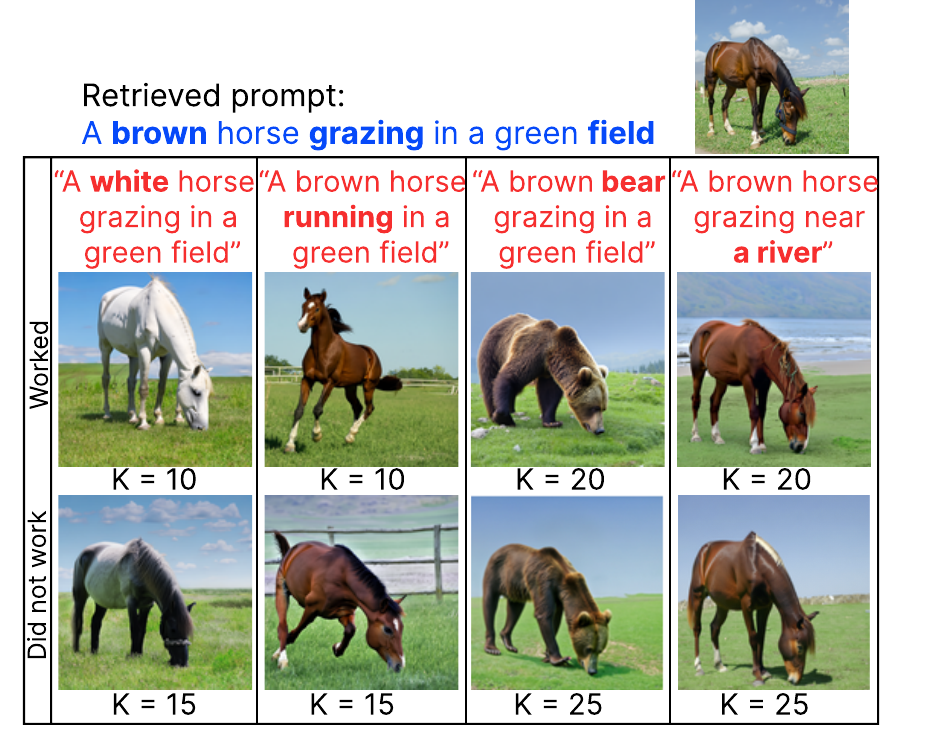}
   \caption{A noise from a brown horse successfully transforms into a white horse at K=10, while K=15 fails. The same noise becomes a bear even at K=20.}
   \label{fig:k_wise}
 \end{minipage}%
 \hfill
 \begin{minipage}{0.31\textwidth}
   \includegraphics[width=1.0\textwidth]{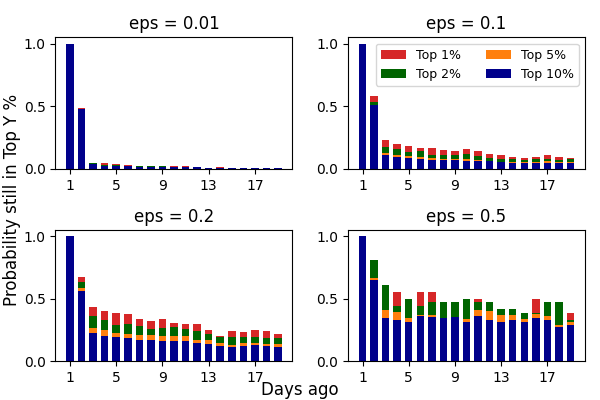}
   \caption{Popularity of top Y\% popular prompt clusters over days for different similarity threshold between prompts ($eps$ value)}
   \label{fig:cluster}
 \end{minipage}%
 \hfill
 \begin{minipage}{0.31\textwidth}
    \includegraphics[width=1.0\textwidth]{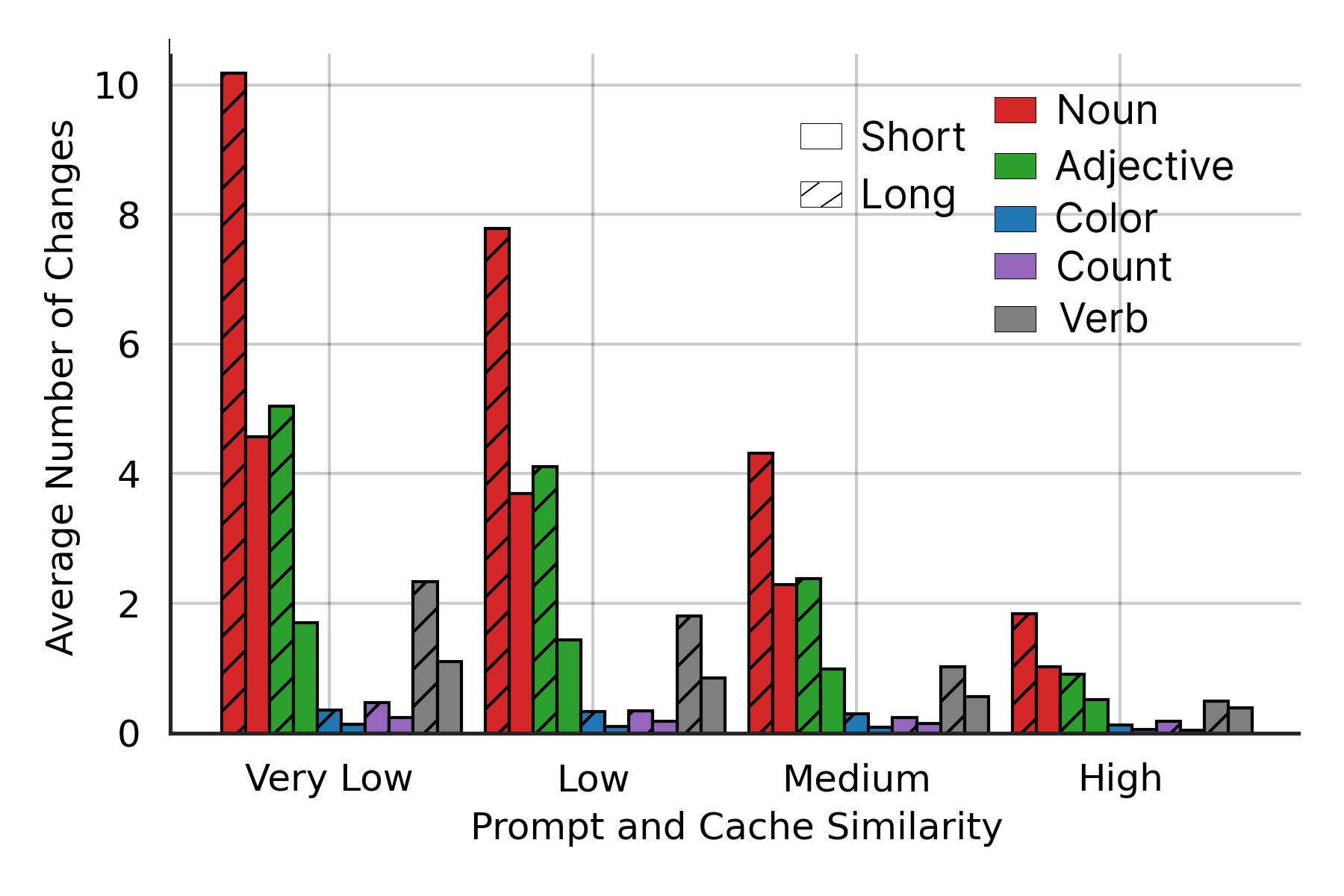}
    \caption{Average number of noun, adjective, colour, count, and verb changes across similarity for both short and long prompts}
    \label{fig:long_short}
  \end{minipage}
\vspace{-1em}
\end{figure*}

The amount of reconditioning needed for the retrieved noise to suit for the new prompt depends on $K$. We observe that various concept/characteristics of an image, like, \textit{layout}, \textit{color}, \textit{shape or objects}, \textit{style}, etc. are not easy to modify beyond certain $K$. With initial noise being random Gaussian, similar to recent works~\cite{zhang2023prospect}, we observed for our dataset that LDM models~\cite{rombach2022high} decide the layout of the objects first within the initial $\sim$20\% of diffusion steps. Color map for the overall image then gets decided within $\sim$40\% steps followed by the shape and the size of the objects and then the artistic style of the image. Overall, we observed after $\sim$50\% of the steps, the concepts get frozen and no further attempts to recondition the intermediate image reflects in the final generated image. For example, when attempted to recondition the image of a brown horse grazing in a green field (Figure \ref{fig:k_wise}, the color can be modified after $K=10$, but not after $K=15$. Similarly, actions, object, and background could be modified only till certain steps and not beyond that.

%% file: tex/analysis.tex
\section{Understanding User Prompts} \label{sec:user_prompt_analysis}
\label{sec:prompt_analysis}

We first show some characterization on real user prompts in DiffusionDB dataset~\cite{wang2022diffusiondb} have 2 million prompts ranging from 8 Aug, 2022 to 22 Aug, 2022.
Each entry has the user-id, prompt, timestamp and the generated image in the dataset.

\noindent\textbf{Top-Y\% most popular prompt clusters over days}

We cluster the prompt queries for each day using DBSCAN algorithm~\cite{schubert2017dbscan} on their CLIP embeddings \cite{radford2021learning} with multiple $eps$ values ($eps$ controls the maximum distance between two samples for one to be considered the neighbor of other). 

We find top-Y\% (with Y = 1, 2, 5 and 10) clusters and extract the most popular prompts on Day 1. We then track how many of these top-Y\% clusters remain in the top-Y\% on the following days, Day 2 to Day 18, and report this fraction as the longevity probability of popular prompt clusters in Figure~\ref{fig:cluster}. We repeat this analysis with different $eps$ values ($eps$ = 0.01, 0.1, 0.2, 0.5) in DBSCAN algorithm.

\noindent\textbf{Takeaway.} From Figure \ref{fig:cluster}, we see that most of the clusters in the top-Y\% prompt clusters do not remain popular in top-Y\% even on Day 2 when the clusters are tightly packed ($eps=0.01$). There is a more significant drop in popularity after 2 days. 
However, as we increase $eps$ and make clusters more loosely packed, it effectively gathers more prompts within a cluster and hence the decay rate of popular prompts is less. We empirically verified that with increasing $eps$, higher number of prompts and more dissimilar prompts are clustered together and hence the longevity of top-Y\% cluster increases. Approximate caching idea works even with less similar prompts, as it can recondition the retrieved noise. Thus, the high longevity of popular cluster as seen in plots for $eps$=0.2 and 0.5 shows potential for approximate caching.   

\noindent\textbf{Most similar short and long prompt pairs}

We divide the prompt queries into short and long prompts, based on the $70^{th}$ percentile word count ($\approx$15 words). 
For each set, we then form pairs of most similar prompts using cosine-similarity between their CLIP embeddings and group them into 4 buckets based on their similarity scores: \textit{Very Low}(less than 0.65), \textit{Low}(0.65 to 0.8), \textit{Medium}(0.8 to 0.9), and \textit{High}(more than 0.9).
For each bucket we analyze what attributes (\eg noun, adjective, verb, color, count) changed between pair of prompts within the same bucket and show the average number of changes along these attributes, for both long and short class of prompts in Figure~\ref{fig:long_short}.

\noindent\textbf{Takeaway.} In Figure \ref{fig:long_short}, we see that as the similarity of the prompts within a pair increases, the average number of changes decreases. We also see that within each bucket, the average number of noun changes are the most, followed by adjective and verb changes. Longer prompts have more changes as compared to the shorter prompts within the same similarity bucket. This indicates that even when several attributes in the text of the prompt changes, the CLIP embedding failed to appropriately distinguish the difference between the two prompts. This highlights a limitation in identifying similar prompts when prompts are very long, as two prompts can be misleadingly retrieved as very similar. 

\begin{figure}[h]
    \centering
    \begin{subfigure}[b]{0.48\linewidth}
        \includegraphics[width=1.0\textwidth]{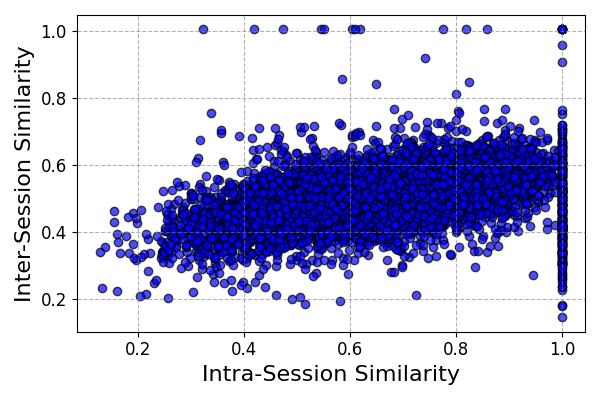}
        \caption{}
        \label{fig:inter-intra}
    \end{subfigure}
    \begin{subfigure}[b]{0.48\linewidth}
        \includegraphics[width=1.0\textwidth]{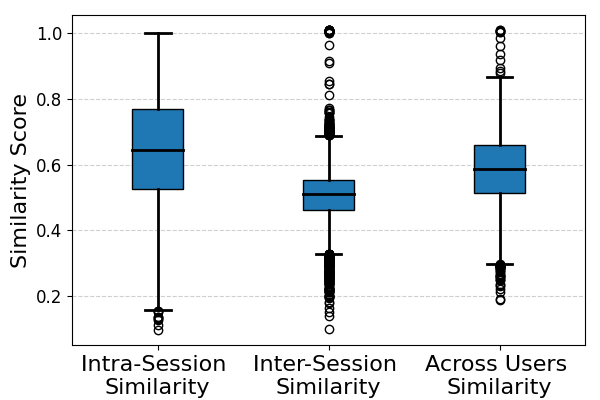}
        \caption{}
        \label{fig:box_plot}
    \end{subfigure}
    \caption{Query prompt similarity (a) for inter v/s intra-session  per user, (b) range for intra, inter per user and across diff users}
    \label{fig:enter-label}
\end{figure}

\noindent\textbf{Intra vs. inter session similarity in prompt queries}

We group the prompts per user, then divide the prompts from that user into 1-hour sessions. We then compare the similarity of prompts within the same sessions (intra) and across different sessions (inter) of the \textit{same user} in Figure \ref{fig:inter-intra}. In Figure \ref{fig:box_plot} we further compare these with similarity scores between prompts from sessions of 100 random \textit{other users}.

\noindent\textbf{Takeaway.} Figure~\ref{fig:inter-intra} shows, many users exhibit high intra-session similarity, meaning users tend to use very similar prompts within a session with lot of repetitions (indicated by a cluster of points at 1). 
Also, there is high inter-session similarity, but it is lower than intra-session, indicating diverse queries across sessions. In Figure \ref{fig:box_plot}, we observe that prompts within a session exhibit high similarity. Notably, prompts across sessions of the same user and even with prompts from different users also demonstrate significant similarity. Interestingly, the similarity between prompts from different users is notably higher, suggesting a promising opportunity for efficient approximate caching utilizing the power of the masses.

%% file: tex/overview.tex
\section{\sys Overview}
\label{sec:overview}

\subsection{Approximate Caching} \label{sec:approximate_caching}

For a new query prompt $\mathcal{P}_Q$, \sys uses approximate-caching to reduce computation by retrieving an intermediate state that was created after $K^{th}$ iteration of a prior image generation process and directly reusing and reconditioning that for $\mathcal{P}_Q$ for the remaining $N-K$ steps.

\textbf{Analytical Modeling:} 
Let $\mathcal{L}$ denote the total end-to-end latency of image generation using approximate caching. Within this, $\mathcal{C}$ represents the cumulative GPU computation time for $N$ diffusion model steps. The set of possible values for $K$ is denoted as $\mathcal{K}$. Each search operation in the vector database (\VDB) incurs a latency cost denoted as $l_s$, and retrieving the intermediate state from the cache introduces a latency denoted as $l_r$. We use $f_c$ to indicate the overall compute savings.

Therefore, for prompts effectively utilizing \textit{approximate caching} with a cache generated at $K$, the total latency experienced can be expressed as:

\begin{equation}
    l_s + \mathcal{C} \cdot \frac{N-K}{N} + l_r
\label{eq:latency}
\end{equation}

In contrast, prompts for which \sys cannot locate a match in the cache will undergo a total latency of 
$l_s + \mathcal{C}$
.

This distinction arises from the design, where it attempts to retrieve an intermediate state from the file system (incurring latency overhead $l_r$) only when a hit is confirmed in the \VDB, ensuring the existence of the state in the cache.

Let us denote the \textit{hit-rate@$K$} for approximate caching as $h(K)$ which is defined as the likelihood that when an intermediate state from $K^{th}$ diffusion step is used, it takes at most $N - K$ diffusion steps to generate a faithful reconditioned image where $N$ is fixed. That is, $(1-h(K))$ fraction of cache exist, which we cannot recondition by running $N-K$ steps.

Now at $K=0$ (running diffusion model from scratch), \textit{all} historical prompts are theoretically usable since image can be reconditioned in at most $N-0$ steps, leading to $h(0) = 1.0$. As $K$ increases, $h(K)$ decreases since we can use only a smaller fraction of intermediate states from $K^{th}$ step to recondition an image by running diffusion at most $N-K$ steps. For lower values of $K$, $h(K)$ is less than 1.0 but can still be relatively high. That is, the diffusion models can effectively recondition the retrieved state if the state is from the initial diffusion steps, resulting in the generation of faithful images.

The decrease in $h(K)$ is influenced by how dissimilar the prompts are. When $K$ surpasses a certain threshold, denoted as {\small $K_T$}, the retrieved state is no longer suitable for further reconditioning, as discussed in \S~\ref{sec:Iteration_analysis}, and thus, { $h(K \ge K_T) = 0$}.

Consequently, the effective fraction of savings in GPU computation for a given $K$ can be expressed as:

{
\begin{equation}
    f_{C} = h(K) \cdot \frac{K}{N}
\label{eq:compute_saving}
\end{equation}
}

It is evident that substantial savings can be achieved when both $K$ and $h(K)$ are sufficiently high. However, the challenge lies in the fact that as $K$ increases, $h(K)$ tends to decrease while aiming to maintain the quality $\mathbb{Q}$ of the generated images\footnote{
For example, if we assume that $h(K)$ decreases linearly from $1.0$ at $K=0$ to $0$ at $K=K_{T}$ following the equation $ h(K) = -\frac{K}{K_{T}} + 1 $, then the optimal single value of $K_{OPT}$ that maximizes fractional savings will be: $K_{OPT} = K_{T}/2$, resulting in effective compute savings of $f_{C}^{max} = K_{T}/4N$.
Similarly, for a slowly decaying quadratic form expressed as
$h(K) = - \left( \frac{K}{K_T}\right)^{2} + 1$,
$f_{C}^{max} = \frac{2 K_{T}}{3\sqrt{3}N} > \frac{K_{T}}{4N}$. Therefore, the slower the decay of $h(K)$ with respect to $K$ the higher the compute savings we can expect.}.
This trend is illustrated in Figure~\ref{fig:choose-k-heuristic} for \DiffusionDB dataset~\cite{wang2022diffusiondb} across different discrete $K$ values.

Now we define $h_{opt}(K)$ as the fraction of cache stored at $K^{th}$ diffusion step that is used to exactly recondition an image for $N-K$ steps. For example, with $N=50, K=5$, we get $h_{opt}(K)$ is the fraction of cache that can be used to recondition an image by running diffusion steps for exact $45$ steps. Thus,

{
\begin{equation}
    h_{opt}(K) = h(K) - h(K'), \text{ where } \argmin_{K'}(K' > K)
\end{equation}
}

{
\begin{equation}
  h(\min\mathcal{K}) = \sum_{K \in \mathcal{K}} h_{opt}(K) 
\label{eq:overall_hit}
\end{equation}
}

In essence, $h_{opt}(K)$ quantifies the probability that $K$ represents the maximum potential savings for incoming prompts. $h(\min \mathcal{K})$ represents the overall hit-rate, \ie fraction of $\mathcal{P}_Q$ having a cache hit.

Our primary objective is to minimize end-to-end latency ($\mathcal{L}$) while maintaining the quality ($\mathbb{Q}$) of generated images. We operate under the constraint that reconditioning of image with cache at the selected $K$ values must ensure a specified level of image quality $\mathbb{Q}$ compared to when image is generated from scratch. The goal is to find the optimal $K$ value that satisfies these objectives and the below quality constraint.

Thus, for a given incoming prompt $\mathcal{P}_Q$ and its corresponding cached prompt $\mathcal{P}_c$

\textbf{Objective} (Minimize $\mathcal{L}$): (following Eq. \ref{eq:latency})

{
\begin{equation}
\min_{K} \mathcal{L} = \sum_{K \in \mathcal{K}} \left(l_s + h_{\text{opt}}(K) \cdot l_r + h_{\text{opt}}(K) \cdot \mathcal{C} \cdot \frac{N - K}{N}\right)
\end{equation}
}

\textbf{Quality Constraint:}

{
\begin{equation}
     \mathbb{Q}(I_K^c | \mathcal{P}_{c_K}, \mathcal{P}_Q) > \alpha \cdot \mathbb{Q}(I_0 | \mathcal{P}_Q) 
\label{eq:quality_constrait}
\end{equation}
}

where $I_K^c$ represents the image generated by using cache $c$ at $K$ and then reconditioning for $N-K$ diffusion steps. $\alpha \in [0,1]$ represents the tolerance threshold over the quality of images generated and is such that $I_K^c$ is not much worse than $I_0$. In our implementation (\S~\ref{sec:cache-selector}), we employ the CLIPScore metric~\cite{hessel2021clipscore} to define $\mathbb{Q}$ with $\alpha = 0.9$. We opt CLIPScore due to its widespread use in evaluating image quality.

In our use case, where $l_r, l_s << \mathcal{C}$, the objective reduces to 

{
\begin{equation}
\min_{K} \mathcal{L} = \sum_{K \in \mathcal{K}} \left( h_{\text{opt}}(K) \cdot \mathcal{C} \cdot \frac{N - K}{N}\right)
\end{equation}
}
which maximizes $K$ and $h_{opt}(K)$ to obtain minimal latency.

This framework demonstrates the relationship between latency and quality in the context of approximate-caching.

Notably, the wasted overhead of \sys can be captured as $(1-h(K)) \cdot l_s$ where the vector database is queried and results in cache miss. This means if we operate in a setting where $h(K)$ is low even if $K$ is high, \sys may not provide latency benefits if $l_s$ is comparable to GPU compute latency $\mathcal{C}$ and such overhead reduction can be an important design aspect as we discuss in \S~\ref{sec:Match_predictor}. 
It is important to note that, due to disparate cost of high-end GPUs, even if there is not much latency reduction, it can still be significantly cost-effective to use \sys as it drastically cuts expensive GPU compute cost on the cloud. 
In our setting we observed $l_s$ is of the order of 100 $ms$ whereas $\mathcal{C}$ is of the order of 10 $s$.

\subsection{System Components}

\begin{figure}[t]
    \centering
    \includegraphics[width=1.0\linewidth]{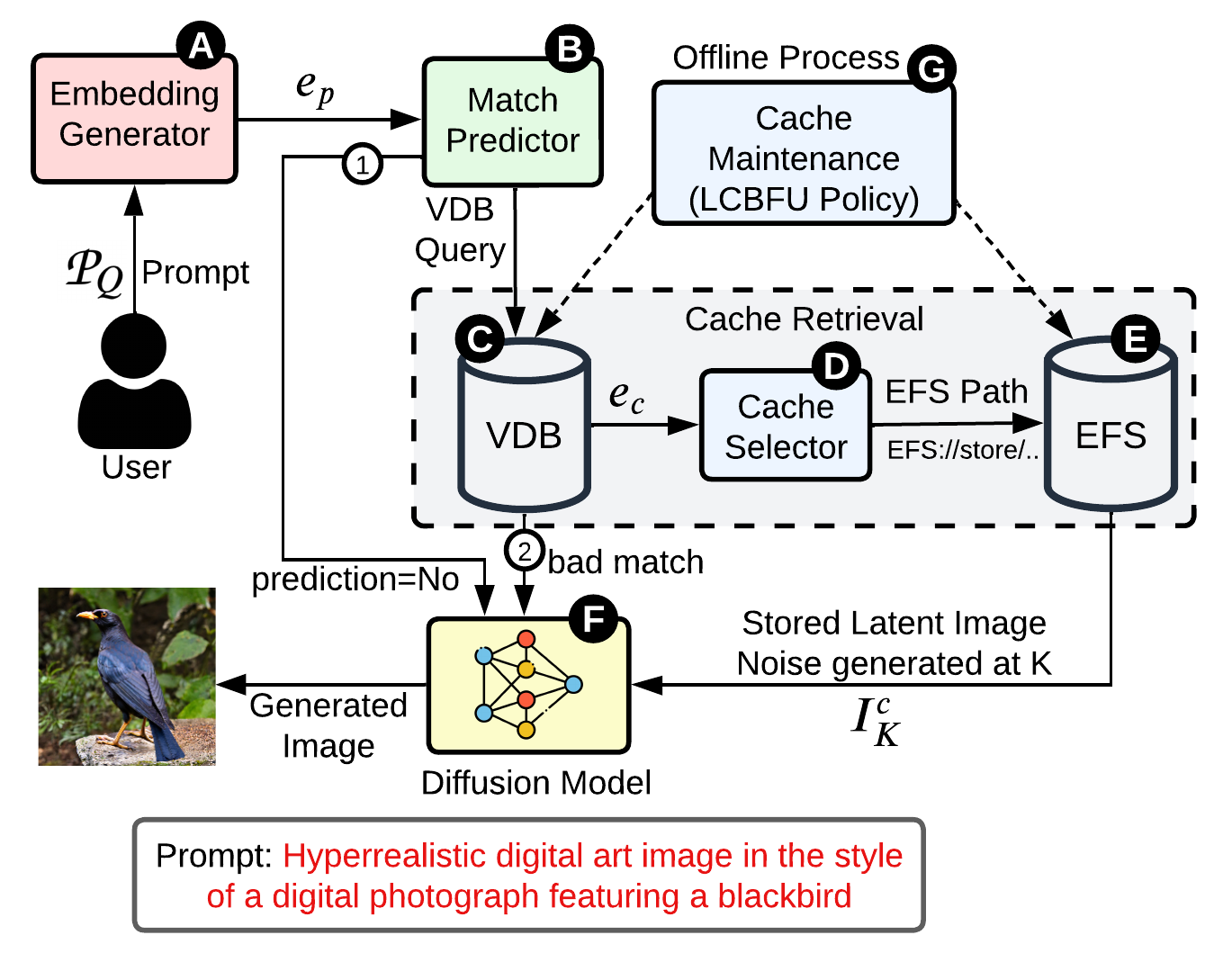}
    \caption{\sys Overview}
    \label{fig:nirvana_overview}
\end{figure}

Figure~\ref{fig:nirvana_overview} shows the main components execution paths of \sys
At first, an embedding vector $e_p$ is generated for the current input prompt $\mathcal{P}_Q$ using an \embeddinggenerator \circled{A}.
Then an optional \matchpredictor module \circled{B} predicts whether there would be a \textit{close enough} match for this embedding in the \textit{vector-database} (\VDB) \circled{C}.

If the presence of a similar enough cached entry is likely, then \sys starts the process of retrieving an intermediate state from the cache.
First a search-query is sent to the \VDB  to find the closest cached embedding of $e_p$, denoted by $e_c$. A VDB uses efficient approximate nearest neighbor (ANN)~\cite{malkov2018efficient} search to find such closest embeddings. 
For each cached historical prompts in the \VDB, \sys stores several intermediate states during the vanilla diffusion process. 
Which of these intermediate states corresponding to the match prompt $e_c$ is optimal for the new prompt vector $e_p$ is calculated using a heuristic by the \cacheselector module \circled{D}. 
Then this particular intermediate state is retrieved from storage \circled{E} (i.e. Elastic File System for \sys) using the pointer of the storage location pointed by the search result of the \VDB query and the particular intermediate state number (i.e. $K$ as explained in \S~\ref{sec:approximate_caching}) calculated by the cache-selector module. An intermediate state $I^{c}_{K}$ is an $L \times D$ dimensional latent representation captured during the denoising process of $e_c$, after $K_{th}$ step. 
Finally, this retrieved intermediate state $I^{c}_{K}$ is passed to the \dm 
\circled{F} along with $e_p$ for it to be reconditioned for image generation for the rest of the denoising steps. 
However, there are two situations when \sys directly falls back to the vanilla diffusion model to generate an image from scratch, sacrificing any optimization: (i) when match-predictor module predicts that a close entry in the \VDB is unlikely (arrow \circledw{1}), and (ii) 
when \VDB query returns a match $e_c$ that is very dissimilar to $e_p$ (arrow \circledw{2}) .

To maintain the cache under a fixed size of  storage and to prevent the \VDB from arbitrarily growing and keep on storing stale entries, a \cachemaintainer module \circled{G} works in the offline mode and implements the novel \policy protocol to keep both the cache and \VDB entries fresh. 

\subsection{\sys Algorithm} \label{sec:nirvana_algo}

Algorithms~\ref{alg:generate_image_cache} shows the overall logic of \sys. First it uses the \matchpredictor to predict if there is a close match for the prompt query $\mathcal{P}_Q$. If the prediction is yes,  it makes a \vdb call to get the nearest prompt and finds the $K$ using heuristics (see \ref{sec:Match_predictor}). Next, it retrieves the intermediate image noise at $K^{th}$ step from  \efs, and passes it to the diffusion model to generate the final image in $N-K$ steps. If there is no match predicted or found, \sys takes fallback to generate an image from scratch.

\begin{algorithm}
\caption{GenerateImageCache($\mathcal{P}_Q$)}
\label{alg:generate_image_cache}
\begin{algorithmic}[1]
\STATE $P_e \gets$ Embed($P$)
\IF{MatchPredictor($P_e$) is $True$}
    \STATE \textcolor{blue}{// Generate using cache}
    \STATE $(\text{neigh}, \text{score}) \gets$ search\_VDB($P_e$)
    \STATE $K \gets$ heuristics\_K($\text{score}$)
    \IF{$K \neq 0$}
        \STATE $path_K \gets \text{neigh}['\text{payload}']['\text{noise}'][K]$
        \STATE $c\_noise \gets$ retrieve\_EFS($path_K$)
        \STATE $I \gets$ model($P, c\_noise, K$)
    \ELSE
        \STATE \textcolor{blue}{// No suitable cache found, generate from scratch}
        \STATE $I \gets$ model($P$, null, 0)
    \ENDIF
\ELSE
    \STATE \textcolor{blue}{// Generate from scratch}
    \STATE $I \gets$ model($P$, null, 0)
\ENDIF
\RETURN $I$
\end{algorithmic}
\end{algorithm}

%% file: tex/design.tex
\section{\sys Design Details}
We now present detailed design of \sys. 

\subsection{Embedding Generator}

Similar to vanilla \dms, \sys first computes a 768-dimensional vector CLIP embedding~\cite{radford2021learning} ($e_p$) from the text prompt.
CLIP effectively positions visually similar prompts closer in the embedding space, which optimizes the likelihood of cache hits in our case.

\subsection{Match Predictor}
\label{sec:Match_predictor}

Using $e_p$, \sys could directly make a search query to \VDB for the closest prompt in the cache. If cache exists, there will be substantial savings in GPU usage for image generation. However, in case of \textit{cache miss}, the search to \VDB becomes a latency overhead without getting any reduction in the GPU computation.
Now, several factors including: a) if it is in the same LAN vs. far away from the GPUs, b) the particular architecture of the \VDB and its internal indexing mechanism, and c) the compute resources dedicated to it etc., dictates the magnitude of the wasted latency overhead during each \textit{miss}. 

To reduce this overhead, \sys uses a component called \matchpredictor, (\circled{B} in Figure~\ref{fig:nirvana_overview}), which \textit{predicts} if an embedding close enough to $e_p$ is \textit{likely} to be present the \VDB. 
If the prediction says it is unlikely, then \sys simply bypasses cache retrieval flow altogether, reducing the wasted overhead. 
Additionally, \matchpredictor also reduces \VDB load corresponding to search misses, improving scalability.

Internally, \matchpredictor uses a light-weight classifier for predictions which runs on the CPU of the same node where the \dm runs on the GPU. This reduces the classification latency by orders of magnitude compared to a query to the \VDB, making latency overhead insignificant. 

Recall from \S~\ref{sec:approximate_caching} that analytically 
the latency overhead is 
{\small $(1-h(K)) \times l_s$}. Now let $c_p$ denote the \textit{precision} of the \matchpredictor classifier. The effective overhead of \sys with an active \matchpredictor is then:

{
\[ 1- max(h(K), c_p) \cdot l_s,
\text{ where } h(K), c_p \in [0,1] \]
}

This means that either when $h(K)=1.0$, \ie prompts are so similar that for every incoming prompt there is a suitable match available in the cache, or when $c_p=1.0$, \ie the classifier is perfect in predictions, the system will not have any wasted latency overhead.   
Note that, when $c_p < 1.0$, \sys would miss some opportunity of compute savings during false-negative cases as it would directly fall back to vanilla \dm and generate an image from scratch instead of attempting to retrieve an intermediate-state. 
For this classifier, \sys uses One-Class Support Vector Machine (One-Class SVM)~\cite{oneclassSVM} which constructs a decision function for outlier detection. This is trained by utilizing all prompt embeddings stored in the \VDB, and assigning them a positive label of 1. To achieve a high precision and recall, we effectively overfit the model to the existing prompt embedding space. Further, we use Stochastic Gradient Descent (SGD) to enable faster retraining of the classifier when embeddings in \VDB changes significantly (\ie > 5\%). 
\sys achieves a $c_p = 0.95$ for real production prompts.

\noindent\textbf{Why not a Bloom Filter?}
Similar to our motivation, Bloom filters~\cite{bloom1970space} are popularly used as a low-cost mechanism to check cache entries in various web services and database systems~\cite{bloomfilter_db}. 
Bloom-filters use hashing algorithms to calculate if an item is likely to be present in the cache with zero false negative rate but with some false positive rate.
However, Bloom filters are not suitable for \sys as we do not search for an \textit{exact match} for cache entries, we search for the \textit{nearest neighbors} in the \VDB by using ANN algorithms.

\subsection{Cache Retrieval}
\label{sec:cacche_retrieval}
Cache Retrieval and \cachemaintainer are the main components of \sys. 
In this section we elaborate the design of the cache retrieval phase which has three internal components \VDB, \cacheselector and storage-system.
In \S~\ref{sec:cache-maintainance} we discuss the design of \cachemaintainer.

\noindent\textbf{Vector Database:}
\sys stores the embedding of the historical prompts in a vector-database (\VDB) for fast and efficient similarity search with the embedding of incoming prompt.
A \vdb uses indexing methods like quantization, graphs, or trees to store and perform high-dimensional similarity search over vectors. 
For a query embedding vector, it can find $m$ approximate nearest neighbors.
In \sys, for each incoming search with $e_p$, \vdb already populated with the embeddings, its payload points to the path where the corresponding intermediate states of $e_c$ at different $K$s are stored.  
\sys uses \texttt{cosine-similarity} as a measure to find the nearest neighbor.
While \sys can work with any \vdb such as \texttt{Qdrant}~\cite{qdrant}, \texttt{Milvus}~\cite{milvus}, \texttt{Weaviate}~\cite{weaviate}, \texttt{Elasticsearch}~\cite{elasticsearch}, in this paper we present results with \texttt{Qdrant} as benchmarking~\cite{vdb_benchmarks} shows low read latency at scale and moderately low update and delete latency.

\noindent\textbf{Elastic File System as Storage:} 
    \sys stores the actual intermediate noise states on AWS Elastic File System (EFS). 
    After comparing with LustreFS~\cite{lustrefs} on read latency, throughput and storage cost, we chose EFS as our storage system.  
    The size of the files containing intermediate-state depends on the architecture of the \dm. For \sys the default \dm uses an intermediate-state of the size of 144 KB and stores for 5 distinct values of $K \in \mathcal{K} = \{5,10, 15, 20, 25\}$.

\subsubsection{Cache Selector} \label{sec:cache-selector}

The \cacheselector component primarily determines which $K \in 
\mathcal{K}$ for $e_c$ from EFS should be retrieved to recondition the intermediate state for the rest of the $N-K$ steps with prompt $\mathcal{P}_Q$ for maximum compute saving while maintaining acceptable image quality.

\begin{figure}[t]
\centering
\begin{minipage}{0.59\columnwidth}
  \includegraphics[width=\textwidth]{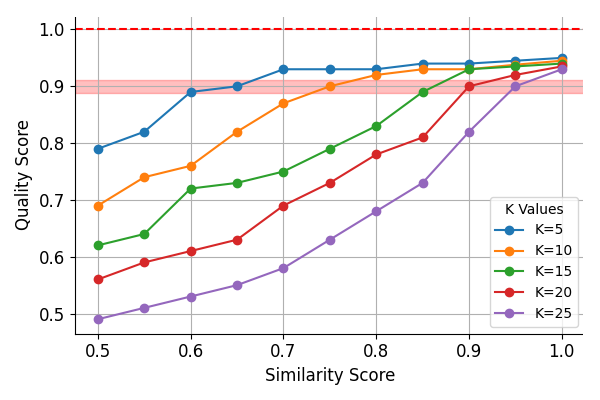}
  \caption{Quality of image generation with cache vs. similarity score across $K$. Thresholds for cache usage at different $K$ values.}
  \label{fig:choose-k-heuristic}
\end{minipage}%
\hfill
\begin{minipage}{0.38\columnwidth}
\begin{lstlisting}[style=mystyle]
def cache_selector(s):
  if s > 0.95: k = 25
  elif s > 0.9: k = 20
  elif s > 0.85: k = 15
  elif s > 0.75: k = 10    
  elif s > 0.65: k = 5
  else: k = 0
  return k
\end{lstlisting}
\caption{Determining $K$ based on similarity score ($s$).}
\label{Fig:code_heuristics}
\end{minipage}
\end{figure}

\noindent\textbf{How do we choose the $K$? }
We observed empirically that the number of steps that can be skipped is correlated to the similarity between $e_p$ and $e_c$. 
Based on this observation, we perform an offline profiling/characterization (Algorithm \ref{alg:heuristic_selection}) to find the appropriate $K$ such that the intermediate-state generated at $K^{th}$ diffusion step is optimal based on the similarity score between the two prompt's embeddings (Eq. \ref{eq:quality_constrait}).
The algorithm ("$Cache Selector$-Profiling(.)"), generates images at each value of $K$ for a set of prompts with their nearest cache prompt. It then finds the minimum similarity score such that all generated images are above a quality threshold $\alpha$. This is then chosen as the minimum similarity at which value of $K$ works and is stored to be used later at run-time.
As shown in Figure \ref{fig:choose-k-heuristic}, we plot the image quality (higher the better) against various similarity scores between the query and the cached prompts across multiple values of $K$.

We use the profiled information to determine the optimal value of $K$ for image generation at run-time. Using the similarity score $s$ between $e_p$ and $e_c$ obtained from \VDB, we choose a $K$ such that the quality score at the plot $(x=K, \text{ } similarity\_score=s)$ is $\geq \alpha$. We choose $\alpha=0.9$. In simple words, we search the inverse map stored by Algorithm \ref{alg:heuristic_selection}.
The red band in Figure \ref{fig:choose-k-heuristic} represents the threshold $\alpha$. Each line in the figure, corresponding to different $K$ values, intersects this red band at the specific similarity scores above which that particular $K$ value is applicable. 
Thus, if similarity score is high, we can skip a greater number of diffusion steps and choose higher $K$ to achieve acceptable image quality. 
But, if similarity score is low, we can skip only a limited number of iterations and hence, a lower $K$. 
Figure \ref{Fig:code_heuristics} shows the actual logic for computing the optimal $K$ given a similarity score $s$ between two prompts that is identified for DiffusionDB dataset using the previously discussed profiling technique.
Based on this, we query the cache storage to obtain $I_K^c$.

\begin{algorithm}[t]
\small
\caption{$Cache Selector$-Profiling($\mathcal{K}$, $I_K^c$, $\alpha$)}
\label{alg:heuristic_selection}
\begin{algorithmic}[1]
\FOR{$K$ in $\mathcal{K}$}
    \STATE $[I_K] \gets$ model($\mathcal{P}_Q, \, I_K^c, \, K$) \text{ } $\forall$ \text{ } $\mathcal{P}_Q$
    \STATE $\text{min\_sim} \gets \min\{\text{sim } s \mid \forall I \in [I_K], \, \text{quality}(I) > \alpha\}$
    \STATE $\text{sim\_K\_map}[K] \gets \text{min\_sim}$
\ENDFOR
\RETURN $\text{sim\_K\_map}$
\end{algorithmic}
\end{algorithm}

Once \sys retrieves the appropriate intermediate image noise from the cache, \dm denoises for $N-K$ steps to generate the final image.

\textbf{Quality vs. performance tradeoff:} 
\sys is designed to extract as much compute savings as possible without degrading the quality of images. 
However, if more aggressive compute savings is needed for certain usecases that are ready to sacrifice some accuracy, \sys can trivially expose a \textit{knob} that can be used to trade-off quality vs. compute savings. The \cacheselector heuristic can be biased to select higher values of $K$. This will provide more compute reduction while letting image quality degrade as \sys will get less steps to recondition the retrieved noise according to the new prompt. At the extreme, for $K=N$, \sys will behave exactly like a pure image retrieval system~\cite{zillizte97:online, MakingSt85:online}.  

\subsection{Cache Maintenance: \policy Policy}
\label{sec:cache-maintainance}

In popular production text-to-image system, there is a large stream of incoming prompts.
The cost associated with total storage used increases and \VDB performance also degrades beyond a certain limit of entries if we keep on storing the cache. 
Therefore, even though \sys can theoretically support infinite cache,
the storage cost and increasing search latency would eventually make it unattractive.
The \cachemaintainer component works in the background to maintain the entries in the cache storage in EFS and in VDB.

To achieve this, \sys uses a novel cache maintenance policy \textit{Least Computationally
Beneficial and Frequently Used} (LCBFU) customized for approximate-caching in \dms.
The fundamental idea is that not all intermediate-states are equally beneficial for overall compute and latency savings. Intermediate-states with high $K$ values can provide huge compute savings, but can be used only when $\mathcal{P}_Q$ has high similarity to one of cached items.
Items stored at low $K$ are usable with a variety of prompts since they can be reconditioned even when the similarity with $\mathcal{P}_Q$ is relatively low, but provides low compute savings.

With \policy, \sys can limit a total of 1 TB cache in EFS without degrading any performance or quality, which on average corresponds to 1.5 million intermediate-states or noises and 300k unique prompt embedding stored in the \VDB.

\noindent\textbf{Drawbacks of traditional cache policies: }
Due to the unique compute model of approximate-caching, the traditional cache policies such as LRU (least recently used), LFU (least frequently used), FIFO (first-in first-out) are not very useful in \sys, since they treat each item in the case homogeneously and only focus on the access patterns or arrival sequence while evicting an item. 
As we have discussed in \S~\ref{sec:background_analytical}, hit-rate alone does not determine efficiency of \sys as items with large $K$ values, even though less frequently accessed can provide significant overall compute reduction.  

\noindent\textbf{\policy: }
We design the \policy that takes into account both access frequency of items as well as potential compute benefit in case of a \textit{hit}. 
It evicts items with least \textit{\policy-score} which for each item $i$ is calculated as: $f_i \times K_i$.
Here $f_i$ is the access frequency of item $i$ and $K_i$ denotes which step of the denoising process this intermediate-state belongs to. Notably, since the \cacheselector heuristic determines $K$ based on similarity of $e_p$ and $e_c$, an aggressive heuristic (high $K$ for low similarity), will force high $f_i$ for high $K$ noises and hence low $K$ noises will be evicted.  For example, an item with $K_i=25$ was accessed 100 times, its \policy-score is 2500, while an item with $K_j=5$ was accessed 200 times, its \policy-score is 1000. Thus, the item $i$ has higher \policy-score since it will provide better compute savings while generating image. 
The complete mechanism of \policy is described as follows: 

\begin{itemize}
    \item \textbf{Insertion:} With every cache miss, we directly insert all the intermediate states generated at diffusion denoising step $K \in \mathcal{K}$ to the cache-storage and the corresponding embedding of the prompt to the \VDB.  
    Thus $|\mathcal{K}|$ are stored in the cache storage per prompt. 
    The insertions are performed without any eviction until we reach the target storage limit. After that every insertion is preceded by an eviction.
    
    \item \textbf{Eviction:} For eviction, \policy maintains a running list of  \textit{\policy-score} in a $K$-min heap, and evict the top-$|\mathcal{K}|$ items from the heap root just before inserting  $|\mathcal{K}|$ intermediate-states for a new prompt. The \policy-score evicts image noises which contribute least to compute savings.
\end{itemize}

With this cache eviction policy, cases can arise for a particular prompt where noises at some $K$s are evicted, while noises at other $K$s are still in the cache. This creates \textbf{\textit{holes}} in the intermediate-states stored.

\textbf{Handling of holes:}
Once eviction policy creates a hole, no straightforward way exists to fill that hole as targeted regeneration of intermediate-state is not possible without running the full diffusion process.
However, the heuristic used by \cacheselector (\S~\ref{sec:cache-selector}) is oblivious to the existence of these holes while determining an appropriate $K$ to be used with the retrieved prompt.
\sys handles this situation by choosing the intermediate-state with the largest value $K$ that is less than or equal to the optimal $K$. This ensures that \sys continues to generate high quality images, albeit with little sacrifice in potential compute savings when it encounters the holes. However, we observed that such cases arise only in \textbf{4-5\%} of the prompts, resulting in imperceptible performance degradation. 

\sys does not actively take a strategy to clean such holes. When for a prompt all the intermediate-states corresponding to all the $|\mathcal{K}|$ values turn in to holes, \policy marks that prompt embedding as dirty and removes it from \vdb as well as corresponding metadata from the storage-cache. 

\cachemaintainer performs both insertions and deletions on \vdb in batches and at the same time the classifier in the \matchpredictor is also retrained with the fresh entries in the \VDB. Recall, as mentioned in \S~\ref{sec:Match_predictor} and \S~\ref{sec:cacche_retrieval}, both \VDB update and classifier updates are fast and takes only around 7.5 and 0.04 seconds respectively for 10k records.

%% file: tex/implementation.tex
\section{Implementation}
\sys is implemented in Python using PyTorch~\cite{paszke2019pytorch}. 

\textbf{Batch Processing: } While deploying, our diffusion model takes up around 80\% of GPU memory with batch size 1, so concurrent batching isn't supported.

\textbf{AITemplate (AIT): }
We leverage AITemplate (AIT)~\cite{ait}, a python framework to accelerate inference serving of the PyTorch-based diffusion model by converting it into CUDA (NVIDIA GPU) / HIP (AMD GPU) C++ code. It provides high performance support during the inference process.

\textbf{System Components: } 
Our system components consist of (i) a classifier that uses $SGDOneClassSVM$~\cite{oneclassSVM} from scikit-learn~\cite{scikit-learn} with $\nu=0.001$, where $\nu$ controls the trade-off between training errors and support vectors, (ii) \VDB hosted in a Docker container on an AWS m5.4xlarge EC2 instance with HNSW indexing of 256 for prompt embedding search~\cite{malkov2018efficient}, (iii) AWS EFS system that offers web and file system access, object storage, and scalability,  (iv) MySQL database to record accesses of cache items, and (v) a text-to-image generator powered by a larger version of stable diffusion-based model with DDIM (Denoising Diffusion Implicit Model) sampler~\cite{song2020denoising}. It generates image in $N=50$ iterations utilizing approximately 8 GB of memory for a batch size of 1.

%% file: tex/evaluation.tex
\section{Evaluation}
\label{sec:eval}

We first evaluate the  overall effectiveness of \sys in terms of quality and in providing significant compute and end-to-end latency savings, throughput, and cost savings for serving on AWS cloud platform. We perform a user study with 60 participants to compare the image generation quality against baselines. We present \policy benefits against common cache-management policy to demonstrate the effectiveness of \sys's internal component design. We also perform and present ablation studies. The key takeaways are:

\begin{itemize}
    \item \sys generates \textbf{high quality} image while \textbf{reducing} both GPU usage and end-to-end latency by up to \textbf{50\%}.
    
    \item \sys \textbf{reduces} all three - \textbf{cost}, \textbf{latency} and \textbf{compute} requirements of \dm by $\sim$\textbf{20\%} on average.
    
    \item With a \textbf{27\% improvement }in system \textbf{throughput}, \sys ensures a  stable user experience with minimal response time variations.
\end{itemize}

\subsection{Methodology} \label{sec:eval_method}

\noindent\textbf{Base Diffusion Model:}
As the base text-to-image model we use a larger Stable Diffusion based model (which we call \textbf{\vanilla}), having approximately 1.5 times the number of parameters as compared to the  2.3B parameter Stable Diffusion XL model \cite{podell2023sdxl}. 
Our model operates within a 96-pixel latent space and performs $N=50$ denoising steps to generate an image in 8.59 seconds, on average, on an A10g GPU. The final image generated is of size $768 \times 768 \times 3$.

\noindent\textbf{Experimental Setup:}
We run the \dm and \embeddinggenerator of \sys on a single NVIDIA A10g GPU (24 GB GPU memory), while the other components are run on a 32 core AMD EPYC 7R32 2.8 GHz CPU with 128 GB CPU memory. The GPU and the CPU machines were attached to a sub-network which included EFS and \VDB. We optimize GPU usage by reusing prompt embeddings with diffusion model embeddings, reducing GPU overhead.

\noindent\textbf{Dataset:} We evaluate on two production datasets.
\begin{itemize}
    \item \textbf{DiffusionDB}~\cite{wang2022diffusiondb}: A total of 2M images for 1.5M unique prompts with a total dataset size of $\approx1.6\ TB$. We filtered the dataset and remove NSFW images
    \item \textbf{\X}: Prompts from a production setup \X spanning over 8 weeks, containing over 7M images for 6.2M unique prompts, with a total size of $\approx5\ TB$.
\end{itemize}
Unless otherwise mentioned, we evaluate the results on \DiffusionDB dataset since \X data is proprietary.

\noindent\textbf{Baselines:} 
We compare two versions of \sys: (1) with \matchpredictor (referred as \sys) and (2) \textit{without} \matchpredictor (referred as \sysomp) against the following baselines:
\begin{itemize}
    \item \textbf{\gpt}: Retrieves image for the closest prompt based on BERT embedding similarity. Otherwise, generates image from scratch \cite{zillizte97:online}
    \item \textbf{\pinecone}: Retrieves image for the closest prompt based on CLIP text embedding similarity. Otherwise, generates image from scratch \cite{MakingSt85:online}
    \item \textbf{\clip (Clip Retrieval System)}: Clip Retrieval System~\cite{rom1504c5:online} is another image retrieval method that uses the embedding of the final image generated by the previous prompts when retrieving the closest image for a given input prompt. 
    \item \textbf{\smallmodel}: A smaller diffusion model \cite{rombach2022high} with 860M parameters, consuming only 33\% of compute/latency compared to \vanilla, generating an image by 50 diffusion steps in 3.05 seconds on average on A10g.
   
\end{itemize}

\noindent\textbf{Workload Generator and Cache Preload:}
The workload generator dispatched prompt in the order in which they arrived, using the dataset's arrival timestamp field to create a stream of prompts. Each prompt query was dispatched to the \sys immediately upon the completion of the preceding query. This query stream began only after the initial 10k prompts were employed to preload the \vdb and \efs with their respective caches for kick-starting the system.

\noindent\textbf{Evaluation Metrics:}
We evaluate \sys on various metrics covering both quality and efficiency aspects.
\begin{enumerate}
    \item \textbf{Quality Metrics:}
    \begin{itemize}
        \item \textbf{FID Score (Fréchet Inception Distance):}  Computes difference between two image datasets and correlates with human visual quality perception~\cite{heusel2017gans}.
        \item \textbf{CLIP Score:} Evaluates the alignment between generated images and their textual prompts~\cite{hessel2021clipscore}
        \item \textbf{PickScore:} A metric designed for predicting user preferences for generated images~\cite{kirstain2023pick}
    \end{itemize}
    \item \textbf{Efficiency Metrics:} To evaluate the efficacy of \sys in providing system efficiency, we report average of 5 runs on the \% savings in GPU usage time, \% reduction in end-to-end latency of image generation, \% increase in throughput as number of image generated per second in a cluster and also amortized dollar-cost per image generation.  
    We also report overall \textit{hit-rate}=$h(min\mathcal{K})$ (Eq. \ref{eq:overall_hit}) for \sys.
    
\end{enumerate}

\subsection{Overall Performance on Quality}
\label{sec:eval_overall}
In this section we evaluate how \sys performs significantly better with respect to image generation quality. 

\textbf{Quantitative Generation Performance}:
Table~\ref{table:main_res} summarizes \sys's improvements in terms of the quantitative image quality metrics.
When compared against the retrieval-based baselines (\S\ref{sec:eval_method}), \sys and \sysomp improves performance significantly as captured by all three metrics~\cite{wang2023exploring}. Retrieval based baselines directly retrieve the image generated from the most similar previous prompt for the query prompt. 
Hence, these methods fail to capture the differences between the incoming prompt and the retrieved prompt since similarity metrics are unable to capture these (\S\ref{sec:user_prompt_analysis}).
Thus, all the retrieval-based baselines incur a significant hit in the quality of the image generated which is of utmost importance in production user-facing text-to-image use cases. Compared to \smallmodel, both \sys and \sysomp perform far superior which shows bigger models retains quality even with \textit{approximate-caching}.

The metrics for \vanilla shows the generated image quality without any kind of approximations. The performance of \sys is very close to \vanilla when compared against CLIP and Pick score for both datasets. This shows that \textit{approximate-caching} does not hurt the overall performance.

We measure FID (lower the better) of baselines against images generated using \vanilla for the same prompts. We also compute FID of \vanilla to indicate the inherent variability of the generated images without any change in base model. To calculate FID for \vanilla, we generated 4 sets of images with 4 different seeds and calculate the FID between these resulting $\Comb{4}{2}$ sets of images. 
As can be seen for both datasets, \sys and \sysomp exhibits much lower FID values than even the internal dissimilarities between different sets of generations by the \vanilla model. This means that images generated by \sys will be indistinguishable from the \vanilla model - which is the design goal. 

\sys performs slightly better than \sysomp due to the presence of \matchpredictor which generates image from scratch during predicted cache miss. However, this comes with a very small hit on \sys's efficiency, compared to \sysomp (see \S\ref{sec:sys_efficieny}).

\begin{table}[t]
  \centering
  \resizebox{0.9\columnwidth}{!}{%
    \ra{1.1}
    \begin{tabular}{@{}lcccc@{}}
      \toprule
       && \multicolumn{3}{c}{\textbf{Quality}} \\
      \cmidrule{3-5} 
      \textbf{Dataset} & \textbf{Models} & $FID\downarrow$ & $CLIP\ Score\uparrow$ &  $PickScore\uparrow$ \\
      \midrule
      \multirow{6}{*}{DiffusionDB} & ${\gpt}$ & 7.98 & 25.84 & 19.04 \\
      & ${\pinecone}$ & 10.92 & 24.83 & 18.92 \\
      & ${\clip}$ & 8.43 & 24.05 & 18.84  \\
      & ${\smallmodel}$& 11.14 & 25.64 &  18.65  \\
      & ${\sysomp}$ & 4.94 & 28.65  & 20.35  \\
      & ${\sys}$ & \textbf{4.68} & \textbf{28.81} &  \textbf{20.41}  \\
      \cmidrule{2-5}
      & ${\vanilla}$ & 6.12-6.92 & 30.28 & 20.86 \\
      \midrule
      \multirow{6}{*}{\textbf{\X}} & ${\gpt}$ & 8.15 & 26.32  & 19.11  \\
      & ${\pinecone}$ & 10.12 & 24.43 & 18.83  \\
      & ${\clip}$ & 8.38 & 23.81 & 18.78 \\
      & ${\smallmodel}$ & 11.35 & 25.91 & 18.92  \\
      & ${\sysomp}$ & 4.48 & 28.94  & 20.31  \\
      & ${\sys}$ & \textbf{4.15} & \textbf{29.12}  & \textbf{20.38}  \\
      \cmidrule{2-5}
      & ${\vanilla}$ & 5.42-6.12 & 30.4 & 20.71  \\
      \bottomrule
    \end{tabular}%
  }
  \caption{We compare \sys against several baselines \gpt, \pinecone, \clip that are pure retrieval-based techniques, a smaller-model and with base vanilla diffusion model. CLIP score computes based on text to image score. All others are image to image comparisons, using \vanilla images as Ground Truth. Classifier has 0.96 Precision, 0.95 Recall.}
  \label{table:main_res}
\end{table}

\begin{figure*}[t]
\centering
\vspace{0pt}
\begin{minipage}[t]{0.35\textwidth}
    \vspace{0pt}
    \begin{minipage}[t]{1.0\textwidth}
         \centering
         \includegraphics[width=0.55\textwidth]{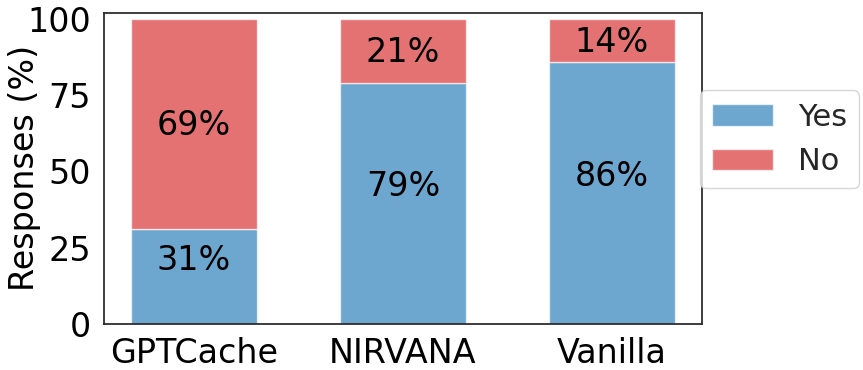}
        \caption{User survey response}
        \label{fig:user_survey_response}
    \end{minipage}
    \begin{minipage}[t]{1.0\textwidth}
    \vspace{2pt}
    \resizebox{1.0\linewidth}{!}{%
    \begin{tabular}{@{}l|c|c|c|ccc@{}}
      \toprule
       &Latency(s) & $90^{th}/median$ & $95^{th}/median$  & $99^{th}/median$ \\
      \midrule
      $GPTCache$ & 2.8 & 29.52 & 29.64  & 29.75  \\
      $\sys$ & 6.9 & 1.20 & 1.21 & 1.21 \\
      $\vanilla$ & 8.6 & 1.01 & 1.02 &  1.02 \\
      \bottomrule
    \end{tabular}%
  }
  \captionof{table}{ Average latency and $n^{th}$ percentile over median values of the response latencies for approaches}
  \label{table:percentile_latency}
  \end{minipage}
\end{minipage}%
  \hfill
  \begin{minipage}[t]{0.31\textwidth}
    \vspace{0pt}
    \includegraphics[width=\textwidth]{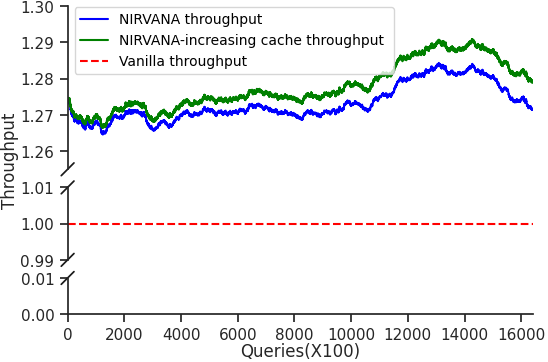}
        \caption{Throughput comparison of models against a stream of queries}
        \label{fig:throughput}
  \end{minipage}%
  \hfill
  \begin{minipage}[t]{0.31\textwidth}
       \vspace{0pt}
        \includegraphics[width=\textwidth]{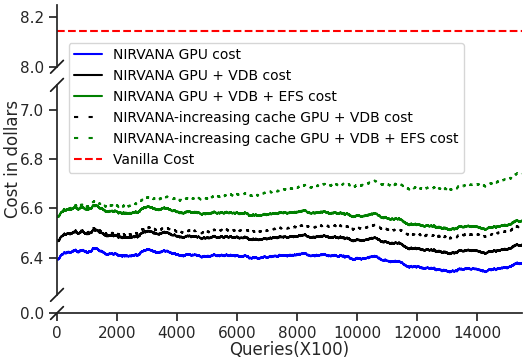}
        \caption{Cost comparison of model components against stream of queries}
        \label{fig:cost}
  \end{minipage}%
  \hfill

\end{figure*}

\textbf{User Study for Accessing Quality}:
We conduct a user study with 60 participants to demonstrate the qualitative analysis. We evaluate around 1000 randomly chosen prompts from \DiffusionDB, where each user was presented with 15 random \texttt{<prompt, generated-image>} pairs and were asked to vote \texttt{Yes}, or \texttt{No} if they feel the generated image properly represents the given prompt. Out of these 15 prompt-image pairs, 
    images of 5 pairs were generated by \gpt, the best performing image-retrieval based baseline, 
    images of next 5 pairs generated by \sys
    and, 5 images using \vanilla.
The questions from these three types were shuffled and presented to each user in a random order and no prompt was repeated within a user session. Users were also given an option to disclose reasons for \texttt{No}.

In Figure~\ref{fig:user_survey_response}, we show the ratio between \texttt{Yes} and \texttt{No} responses for each of the three models aggregated across all users. It can be seen that \gpt gets the lowest number of \texttt{Yes} votes, while \sys is just marginally beneath the upper bound \vanilla. When analyzed for reasons of \texttt{No} ($\sim$12\% of the negative responses), we discovered that 80\% of these reasons were for the images generated by \gpt. These insights underscore the challenges of relying solely on retrieval-based solutions for image generation and hence we do not use these baselines in further evaluation.

\textit{This reinforces the fact that \sys is much superior in maintaining image quality, which is of paramount importance for commercial deployment.}

\subsection{System Efficiency of \sys} \label{sec:sys_efficieny}

We compare \sys in terms of image generation latency, hit rate, compute savings and cost of running \sys.

\textbf{Latency}:
Latency in Figure \ref{table:percentile_latency} refers to the average end-to-end latency across two million prompt queries tested. We plot the $n^{th}$ percentile over median of response latencies for some baselines. An image generation system should exhibit low latency for enhanced user experience. However, pure retrieval techniques like \gpt provide low latency but produce low-quality images and experiences significant fluctuations in their end-to-end latency since it needs to run diffusion model from scratch for certain prompts which are dissimilar from the cached prompts, thus impacting the user experience~\cite{bai2017understanding} as shown in Table \ref{table:percentile_latency}. \sys reduces such variance in latency compared to the baselines and also the overall latency compared to \vanilla providing a much more stable and faster user experience. The key to minimizing response time variability lies in the smoother transition in compute time across different prompts, attributable to \sys's ability to retrieve caches at various values of $K$.

\textit{This observation underscores that \sys not only reduces overall latency by 19.8\% but also minimizes variance in response times across different prompts, ultimately providing a consistent and stable user experience~\cite{broadwell2004response}.}

\textbf{Hit-Rate and Compute Savings}:
Figure~\ref{fig:k_wise_hit} highlights the \textit{hit-rate} and \textit{compute-savings} of \sys across $K$ (x-axis). Averaging over all $K$s, \sys achieves a substantial \textit{hit-rate} of 88\% and noteworthy \textit{compute savings} of 21\% as indicated by the blue and black dotted lines. The cumulative \textit{hit-rate} curve illustrates how the \textit{hit-rate} varies with different values of $K$ and plots $h(K)$. Additionally, the bar plot showcases the potential savings achievable at each $K$ and the actual savings realized at specific $K$ values. For instance, at $K = 25$, there is a potential savings of 50\%, while the actual \textit{hit-rate} is 8\%, resulting in an actual savings of 4\%.

\textbf{Throughput}: We quantify throughput as the number of prompts processed per minute by the system. To evaluate this, we replay the stream of prompts from the DiffusionDB dataset. We assess the relative throughput of \sys in comparison to \vanilla, considering two cache settings. The first setting employs \policy with a cache size of 1.5 million items, while the second setting involves an increasing cache configuration where no cache eviction occurs, effectively providing a theoretically infinite cache size.
Figure \ref{fig:throughput} represents our findings, where the x-axis corresponds to the stream of queries, and we plot the relative throughput of the system. Notably, our results indicate that \sys achieves $\sim$1.28 times higher throughput than \vanilla for both the settings.

\textbf{Cost of Image Generation}: 
To provide a detailed cost breakdown for \sys and \vanilla, we considered specific AWS components and their associated expenses, as per AWS pricing\cite{AWSProdu62:online}.
We use the g5.24xlarge GPU (96 GB GPU) instance from US East region which costs \$8.144 per hour. This cost was used to estimate the GPU-related expenses incurred by \vanilla and \sys. \vanilla solely uses the GPU resources for each image generation.
However \sys also relies on additional components besides GPU, namely \VDB\cite{VectorSe4:online} and EFS.
For \VDB, the cost is estimated at \$0.12 per hour which covers storage and search operations performed over the prompt embeddings.
Additionally, \sys incurs costs associated with EFS. In the US East region, utilizing the standard storage type with 20\% frequent access over elastic throughput cost amounts to \$0.09 per hour.
To determine the overall amortized cost of \sys, we calculated it by dividing the GPU cost by the throughput and then adding the costs related to \VDB and EFS. 
It is important to note that this cost analysis was conducted under identical settings for both systems. Despite the inclusion of these additional expenses, Figure~\ref{fig:cost} highlights that \sys manages to achieve a remarkable 19\% reduction in cost compared to \vanilla. This significant cost efficiency underscores the practical advantages of \sys in real-world deployment.

\subsection{\policy Performance}
\label{sec:eval_cache_policy}

\begin{figure*}[t]
\centering
\vspace{0pt}
  \begin{minipage}[t]{0.37\textwidth}
    \vspace{0pt}
   \centering
   \begin{subtable}[b]{\linewidth}
    \centering
    \ra{1.1}
    \resizebox{0.78\linewidth}{!}{%
      \begin{tabular}{@{}l|c|c|c|c|cc@{}}
        \toprule
        Cache Size(GB) & \#noises in cache & $FIFO$ & $LRU$ & $LFU$ & ${\policy}$ \\
        \midrule
        $1 GB$ & $1500$ & 0.58 & 0.65 & 0.64 & 0.65 \\
        $10 GB$ & $15000$ & 0.68 & 0.77 & 0.78 & 0.77 \\
        $100 GB$ & $150000$ & 0.74 & 0.85 & 0.83 & 0.82 \\
        $1000 GB$ & $1500000$ & 0.83 & 0.95 & 0.94 & 0.93 \\
        \bottomrule
      \end{tabular}%
    }
    \caption{Hit rate}
  \end{subtable}
  \begin{subtable}[b]{\linewidth}
    \centering
    \ra{1.1}
    \resizebox{0.8\linewidth}{!}{%
  \begin{tabular}{@{}l|c|c|c|c|>{\columncolor{lightgray}}c@{}}
    \toprule
    Cache Size(GB) & \#noises in cache & $FIFO$ & $LRU$ & $LFU$ & \cellcolor{white}${\policy}$ \\
    \midrule
    $1 GB$ & $1500$ & 0.11 & 0.12 & 0.12 & 0.12 \\
    $10 GB$ & $15000$ & 0.13 & 0.14 & 0.14 & 0.15 \\
    $100 GB$ & $150000$ & 0.14 & 0.16 & 0.16 & \textbf{0.18} \\
    $1000 GB$ & $1500000$ & 0.17 & 0.20 & 0.19 & \textbf{0.23} \\
    \bottomrule
   \end{tabular}
     }
 \caption{\% Compute savings}
 \label{table:compute_clfu}
 \end{subtable}
 \captionof{table}{Performance of different eviction techniques. Compute savings with \policy eviction is significant.}
 \label{table:compute_hit_evict}
 \end{minipage}%
 \hfill
 \begin{minipage}[t]{0.35\textwidth}
    \vspace{0pt}
    \includegraphics[width=1.0\textwidth]{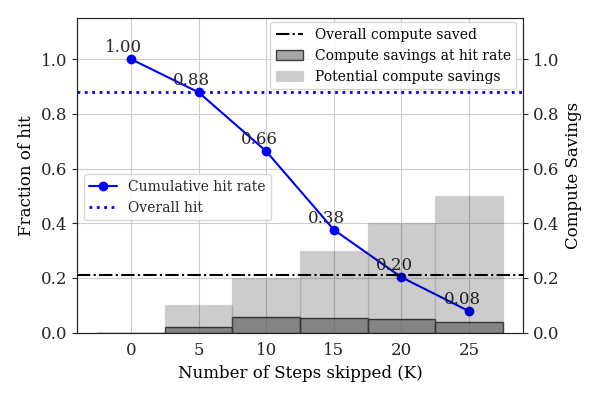}
  \caption{Hit rate and compute saved across K's}
  \label{fig:k_wise_hit}
  \end{minipage}%
 \hfill   
 \begin{minipage}[t]{0.28\textwidth}
    \vspace{0pt}
     \begin{minipage}[t]{1.0\textwidth}
          \centering
        \includegraphics[width=1.0\textwidth]{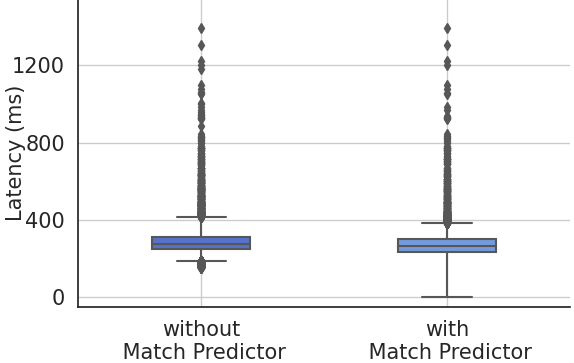}
    \caption{Latency distribution with and without \matchpredictor. }
    \label{fig:latency_class}
    \end{minipage}
    \begin{minipage}[t]{1.0\textwidth}

    \end{minipage}
  \end{minipage}

\end{figure*}

 We experiment the effectiveness of \policy against common caching techniques like FIFO, LRU and LFU. 
 To ensure a fair evaluation, we maintained the same workload generator settings across all experiments with cold cache setting. FIFO removed the earliest added noises from the cache. LRU and LFU evicts cache items based on their access frequency and recency.
 Table~\ref{table:compute_hit_evict} reports \textit{hit-rate} (Eq. \ref{eq:overall_hit}) and \% compute savings (\% of GPU time savings when compared against \vanilla) for various cache eviction policies across different cache sizes. Hit rate of \policy is comparable to LRU and LFU, while outperforming FIFO. Notably, as highlighted in Table \ref{table:table:compute_clfu}(b), the proposed \policy offers substantial compute savings compared to all other policies since it is designed to incorporate $K$ with image access frequency for eviction. \policy is not designed to have the best hit rate.

\subsection{Decomposition of End-to-End Latency in \sys}

\begin{figure}[h!]
\centering
  \includegraphics[width=1.0\columnwidth]{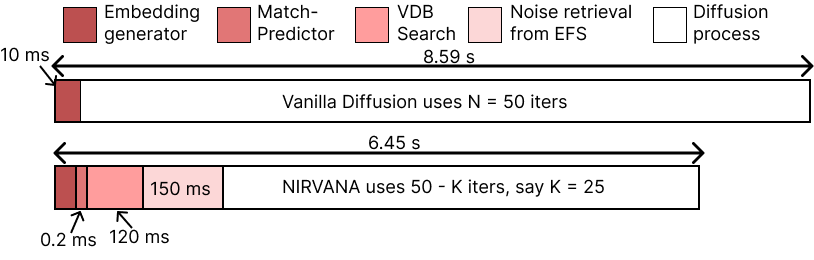}
  \caption{Time taken by different components of \vanilla and \sys for generating image using 50 steps on A10g instance.
  \label{fig:timeline}
  }
\end{figure}

In Figure~\ref{fig:timeline} we show the end-to-end latency of \vanilla diffusion model and also how different components of \sys contribute towards its end-to-end latency. We can see noise retrieval from EFS and VDB search are the main contributors as the overhead. 

\subsection{Image Quality across Long vs. Short Prompts}

As we discussed in \S\ref{sec:prompt_analysis}, the prompt queries can be either long or short. We perform ablation to see how the system works with them. The ablation results presented in Table \ref{table:prompt_len} indicate that \sys performs more effectively with slightly shorter prompts compared to very lengthy ones. This disparity in performance can be intuitively attributed to the retrieval technique employed. Longer prompts tend to challenge the ability of embeddings (in our case, CLIP) to capture the context adequately. If the prompt retrieved from the cache significantly deviates from the query despite high similarity based on $CLIPText$ embeddings (as discussed in \S\ref{sec:prompt_analysis}), it may result in the generation of incoherent images

\begin{table}[h!]
  \centering
  \ra{1.1}
  \resizebox{0.45\linewidth}{!}{%
    \begin{tabular}{@{}lccc@{}}
      \toprule
      Prompt & $FID\downarrow$ & $CLIPScore\uparrow$  \\
      \midrule
      $Short$ & 7.96 & 28.48 \\
      $Long$ & 11.48 & 28.96 \\
    
      \bottomrule
    \end{tabular}%
  }
  \caption{Generation of short v/s long prompts. Less than 15 words is considered short.}
  \label{table:prompt_len}
\end{table}

\subsection{Sensitivity Analysis}
\label{sec:eval_sensitivity}
We now present some sensitivity analysis regrading \sys's design choices.

\textbf{Effectiveness of Match-Predictor:}
%
We conducted an evaluation to assess the effectiveness of using the \matchpredictor within \sys, where we measured the average image generation latency. Figure~\ref{fig:latency_class} shows that \matchpredictor contributes significantly to lowering tail latency values. Furthermore, a considerable fraction of queries has negligible overhead, approximately equal to zero. This behavior is attributed to the \matchpredictor's ability to promptly predict whether a particular prompt is present in the cache or not. It reduces the requirement for I/O-related activities, resulting in decreased latencies. 

\begin{figure}[h!]
\centering
  \includegraphics[width=0.6\columnwidth]{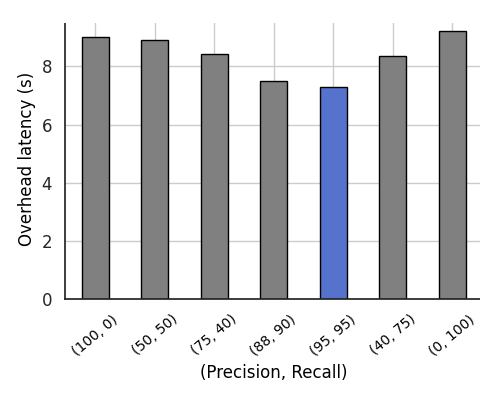}
  \caption{Hit rate and compute saved across K's}
  \label{fig:abl_class}
\end{figure}

\textbf{Setting for Match-Predictor:}
The $SGDOneClassSVM$ \matchpredictor can produce binary predictions (0 or 1) by employing various thresholds. These thresholds influence the Precision (P) and Recall (R) values obtained from the \matchpredictor. To determine the optimal settings, we conducted an ablation study, measuring the overhead latency under different P and R configurations. The resulting plot in Figure \ref{fig:abl_class} led us to select the settings with a $P = 96$ and  $R = 95$

\textbf{Embedding type:}
We conducted a comparison of \sys using various types of embeddings for query and/or cache. Some embeddings are applied to the query as well as cache prompts, while other is applied just to the cached prompt and CLIP embedding is maintained for query prompt. This is indicated by the column `Prompt/Cache'. The results, summarized in Table \ref{table:Ablation_embed}, indicate that the quality of image generation, assessed through metrics such as FID, CLIP Score, and PickScore, remains consistent across different embedding types. However, the hit rate and compute savings achieved with the CLIP Text embeddings significantly outperform the other two embedding types. Hence, we selected CLIP Text embeddings for our system design.

\begin{table}[h!]
  \centering
  \ra{1.1}
  \resizebox{0.9\linewidth}{!}{%
    \begin{tabular}{@{}l|c|c|c|c|cc@{}}
      \toprule
      Embedding & Prompt/Cache & FID$\downarrow$ & CLIPScore$\uparrow$ & HitRate$\uparrow$ & ComputeSavings$\uparrow$ \\
      \midrule
      $BERT$  & Query \& Cache  & 4.53 & 28.94 & 0.77 & 0.16 \\ 
     $CLIPImage$ & Only Cache & 4.85 & 28.49 &  0.84 & 0.18 \\ 
      $CLIP$ & Query \& Cache & 4.94 &  28.65 &  \textbf{0.93} & \textbf{0.23 } \\ 
      \bottomrule
    \end{tabular}%
  }
  \caption{Quality of generation for different embeddings.}
  \label{table:Ablation_embed}
\end{table}

\subsection{Quality with different Caching Policy}
We conducted an evaluation to compare image quality metrics, including $FID$, $CLIPScore$, and $PickScore$, while using different cache eviction techniques. The results, presented in Table \ref{table:quality_evict}, demonstrate that image generation quality remains consistent across all caching techniques. Therefore, the choice of eviction mechanism should prioritize improved compute savings and hit rates.

\begin{table}[h!]
\centering
        \ra{1.1}
  \resizebox{0.6\linewidth}{!}{%
    \begin{tabular}{@{}l|c|c|ccc@{}}
      \toprule
      Policy & $FID\downarrow$ & $CLIPScore\uparrow$  & $PickScore\uparrow$ \\
      \midrule
      $FIFO$ & 5.12 & 28.25 & 20.31 \\
      $LRU$ & 4.82 & 28.54 & 20.38 \\
      $LFU$ & 4.98 & 28.61 &  20.42 \\
      ${\policy}$ & 4.94 & 28.65 &  20.41 \\
      \bottomrule
    \end{tabular}%
  }
  \caption{Quality of generation for different eviction techniques with 1500 GB cache.}
  \label{table:quality_evict}
  \end{table}

%% file: tex/related_work.tex
\section{Related Works}

\textbf{ML optimizations.} Various optimization techniques like model distillation~\cite{meng2023distillation, salimans2021progressive}, pruning~\cite{fang2023structural}, quantizations~\cite{li2023qdiffusion} and others~\cite{lai2023modelkeeper, yin2021towards, hu2021lora} exists for large diffusion models but often degrades the quality as well. DeepSpeed~\cite{deepspeed} is an optimization library for distributed inference and implements multiple techniques for the same. These can make the base model faster but are orthogonal to \sys as they do not fundamentally change the nature of the iterative denosining process that \sys exploits for compute reduction. 

\textbf{Model-serving in Cloud.} 
Past research~\cite{yadwadkar2019case, zhang2019mark, crankshaw2017clipper, gunasekaran2022cocktail, lee2018pretzel} explored efficient ML model serving to reduce inference latency.  Clipper~\cite{crankshaw2017clipper} implements optimizations like layer caching, ensembling methods and straggler mitigation. Cocktail~\cite{gunasekaran2022cocktail} designs a cost effective ensemble based model serving framework along with proactive autoscaling for resource management to provide high throughout and low latency. Other works~\cite{crankshaw2020inferline, zhang2019mark, chen2023symphony} optimized the prediction pipeline cost during load variations. However, these are complimentary to \sys and can be integrated easily.
Tabi~\cite{wang2023tabi} uses multiple heterogeneous  models for Large Language Models. 

\textbf{Text-to-image models.} Diffusion model is one of the main classes of text-to-image models, which was popularized by Dall-E~\cite{dall-e}, Imagen~\cite{saharia2022photorealistic} and Stable Diffusion~\cite{rombach2022high}. Several other enterprises like Midjourney~\cite{midjourney}, Deci~\cite{deci_diffusion} and Adobe~\cite{firefly} have built their own diffusion models. Algorithmic optimizations include designing of faster sampling step~\cite{wu2023fast, kong2021fast, zheng2023fast, song2020denoising, ho2020denoising} and parallel sampling~\cite{shih2023parallel, zheng2023fast} and hence trading off compute for speed. However, our work is orthogonal and the underlying diffusion model can be chosen from any of the above-mentioned works.

\textbf{Caching.} Caching in DNN inference has been explored in the past~\cite{grave2017improving, kumar2019accelerating, crankshaw2017clipper, kuhn-1988-speech} including caching intermediate layer outputs to avoid running every layers again on different input~\cite{kumar2019accelerating, crankshaw2017clipper}. Kumar \etal~\cite{kumar2019accelerating} coins the term \textit{approximate caching} for above, but our semantics is orthogonal since we cache the intermediate image noise, and not the model layer outputs. None of the caching policies implemented in these works (LRU, static cache) work in our case. Other caching techniques~\cite{beckmann2018lhd, yang2021segcache} are non-trivial to extend for our purpose. Retrieval-based works~\cite{rom1504c5:online, MakingSt85:online, zillizte97:online} uses caching to retrieve image for the most similar prompt, but suffer in quality (see \S\ref{sec:eval}).

\section{Discussions}
\label{sec:discussions}

\textbf{Image diversity.}
Since \sys reuses previous intermediate-states from the cache, over time the diversity of images generated by the system can reduce if majority of the prompts encountered by the system are very similar. 
This can be addressed in two ways: first, by actively changing the \texttt{seed} of the denoising process after retrieval. We observed that for $K< 35$ this can increase diversity. If prompts become too similar, then \sys would attempt to save compute more aggressively using a higher $K$ where \texttt{seed} change becomes ineffective. In that case \sys can be designed to let ab $\epsilon$ fraction of prompts follows the vanilla diffusion process to maintain diversity of images. 

\textbf{Change in prompt characteristics.}
If the characteristics of the prompts suddenly change due to some external factors, \sys will find less similar items in the cache and will automatically move towards lower $K$ or even for more dissimilar prompts the \matchpredictor will kick in and redirect the prompts to vanilla diffusion process. While this will reduce compute savings, \textit{by design} \sys will not let the image quality degrade in case of a sudden change in prompt characteristics.    

\textbf{ML optimizations.} 
Several techniques are used to reduce the compute footprint and latency of models such as use of lower precision~\cite{li2023qdiffusion}, distillations~\cite{meng2023distillation, salimans2021progressive}, pruning~\cite{fang2023structural}, batched-inference optimizations~\cite{batched_inference}. 
\sys is complementary to such techniques as it can be used on top of those optimized model as well to reduce redundant computation using approximate-caching.
However, a new family of generative models emerge that does not require such large number of iterative steps, then \sys's applicability will become limited. But since as of today, \dms with 50 or more denoising steps produce the best and production quality images, \sys provides an attractive proposition for compute reduction.

%% file: tex/conclusion.tex
\section{Conclusion}
In this paper, we introduced the design and implementation of \sys that uses a novel technique called \textit{approximate-caching} to significantly reduce compute cost and latency during text-to-image generation using diffusion models by caching and reusing intermediate-states created while processing prior text prompts. We also presented a new cache management technique to manage these intermediate states to optimize performance under a fixed storage limit.

%% file: main.bbl
\begin{thebibliography}{10}

\bibitem{AWSProdu62:online}
Aws product and service pricing | amazon web services.
\newblock \url{https://aws.amazon.com/pricing/?aws-products-pricing.sort-by=item.additionalFields.productNameLowercase&aws-products-pricing.sort-order=asc&awsf.Free%20Tier%20Type=*all&awsf.tech-category=*all}.
\newblock (Accessed on 09/20/2023).

\bibitem{Complete85:online}
Complete guide to samplers in stable diffusion - félix sanz.
\newblock \url{https://www.felixsanz.dev/articles/complete-guide-to-samplers-in-stable-diffusion}.
\newblock (Accessed on 09/21/2023).

\bibitem{vdb_benchmarks}
Hugging face model repository.
\newblock \url{https://huggingface.co/models}.

\bibitem{MakingSt85:online}
Making stable diffusion faster with intelligent caching | pinecone.
\newblock \url{https://www.pinecone.io/learn/faster-stable-diffusion/}.
\newblock (Accessed on 09/21/2023).

\bibitem{rom1504c5:online}
rom1504/clip-retrieval: Easily compute clip embeddings and build a clip retrieval system with them.
\newblock \url{https://github.com/rom1504/clip-retrieval}.
\newblock (Accessed on 09/21/2023).

\bibitem{VectorSe4:online}
Vector search database | qdrant cloud.
\newblock \url{https://cloud.qdrant.io/calculator}.
\newblock (Accessed on 09/20/2023).

\bibitem{zillizte97:online}
zilliztech/gptcache: Semantic cache for llms. fully integrated with langchain and llama\_index.
\newblock \url{https://github.com/zilliztech/GPTCache}.
\newblock (Accessed on 09/21/2023).

\bibitem{adobe_express_usage}
Adobe express with ai-powered firefly integration now commercially available.
\newblock \url{https://news.adobe.com/news/news-details/2023/Adobe-Express-With-AI-Powered-Firefly-Integration-Now-Commercially-Available/default.aspx}, 2023.

\bibitem{firefly}
Adobe firefly. \href{https://www.adobe.com/sensei/generative-ai/firefly.html}{https://www.adobe.com/sensei/generative-ai/firefly.html}.
\newblock 2023.

\bibitem{firefly_news}
Adobe unleashes new era of creativity for all with the commercial release of generative ai.
\newblock \url{https://news.adobe.com/news/news-details/2023/Adobe-Unleashes-New-Era-of-Creativity-for-All-With-the-Commercial-Release-of-Generative-AI/default.aspx}, 2023.

\bibitem{ait}
Aitemplate.
\newblock \url{https://github.com/facebookincubator/AITemplate }, 2023.

\bibitem{dall-e}
Dall-e 2. \href{https://openai.com/dall-e-2}{https://openai.com/dall-e-2}.
\newblock 2023.

\bibitem{elasticsearch}
Elasticsearch.
\newblock \url{https://www.elastic.co/}, 2023.

\bibitem{intel_lab}
Intel labs introduces ai diffusion model, generates 360-degree images from text prompts.
\newblock \url{https://www.businesswire.com/news/home/20230621842353/en/Intel-Labs-Introduces-AI-Diffusion-Model-Generates-360-Degree-Images-from-Text-Prompts}, 2023.

\bibitem{deci_diffusion}
Introducing decidiffusion 1.0: : 3x the speed of stable diffusion with the same quality. \href{https://deci.ai/blog/decidiffusion-1-0-3x-faster-than-stable-diffusion-same-quality/}{https://deci.ai/blog/decidiffusion-1-0-3x-faster-than-stable-diffusion-same-quality/}.
\newblock 2023.

\bibitem{lustrefs}
Lustrefs.
\newblock \url{https://www.lustre.org/}, 2023.

\bibitem{midjourney}
Midjourney. \href{https://www.midjourney.com/home/}{https://www.midjourney.com/home/}.
\newblock 2023.

\bibitem{milvus}
Milvus - vector database.
\newblock \url{https://milvus.io/}, 2023.

\bibitem{bloomfilter_db}
Myrocks and bloom filters.
\newblock \url{https://mariadb.com/kb/en/myrocks-and-bloom-filters/}, 2023.

\bibitem{oneclassSVM}
Oneclass svm.
\newblock \url{https://scikit-learn.org/stable/modules/generated/sklearn.svm.OneClassSVM.html}, 2023.

\bibitem{qdrant}
Qdrant - vector database.
\newblock \url{https://qdrant.tech/}, 2023.

\bibitem{batched_inference}
Stable diffusion batch prediction with ray data.
\newblock \url{https://docs.ray.io/en/latest/data/examples/stablediffusion_batch_prediction.html}, 2023.

\bibitem{weaviate}
Weaviate - vector database.
\newblock \url{https://weaviate.io/}, 2023.

\bibitem{bai2017understanding}
Xiao Bai, Ioannis Arapakis, B~Barla Cambazoglu, and Ana Freire.
\newblock Understanding and leveraging the impact of response latency on user behaviour in web search.
\newblock {\em ACM Transactions on Information Systems (TOIS)}, 36(2):1--42, 2017.

\bibitem{beckmann2018lhd}
Nathan Beckmann, Haoxian Chen, and Asaf Cidon.
\newblock $\{$LHD$\}$: Improving cache hit rate by maximizing hit density.
\newblock In {\em 15th USENIX Symposium on Networked Systems Design and Implementation (NSDI 18)}, pages 389--403, 2018.

\bibitem{bloom1970space}
Burton~H Bloom.
\newblock Space/time trade-offs in hash coding with allowable errors.
\newblock {\em Communications of the ACM}, 13(7):422--426, 1970.

\bibitem{broadwell2004response}
Peter~M Broadwell.
\newblock {\em Response time as a performability metric for online services}.
\newblock Computer Science Division, University of California, 2004.

\bibitem{chen2023symphony}
Lequn Chen, Weixin Deng, Anirudh Canumalla, Yu~Xin, Matthai Philipose, and Arvind Krishnamurthy.
\newblock Symphony: Optimized model serving using centralized orchestration.
\newblock {\em arXiv preprint arXiv:2308.07470}, 2023.

\bibitem{crankshaw2020inferline}
Daniel Crankshaw, Gur-Eyal Sela, Corey Zumar, Xiangxi Mo, Joseph~E. Gonzalez, Ion Stoica, and Alexey Tumanov.
\newblock Inferline: Ml prediction pipeline provisioning and management for tight latency objectives, 2020.

\bibitem{crankshaw2017clipper}
Daniel Crankshaw, Xin Wang, Guilio Zhou, Michael~J Franklin, Joseph~E Gonzalez, and Ion Stoica.
\newblock Clipper: A $\{$Low-Latency$\}$ online prediction serving system.
\newblock In {\em 14th USENIX Symposium on Networked Systems Design and Implementation (NSDI 17)}, pages 613--627, 2017.

\bibitem{croitoru2023diffusion}
Florinel-Alin Croitoru, Vlad Hondru, Radu~Tudor Ionescu, and Mubarak Shah.
\newblock Diffusion models in vision: A survey.
\newblock {\em IEEE Transactions on Pattern Analysis and Machine Intelligence}, 2023.

\bibitem{deepspeed}
deepspeed.ai.
\newblock Deepspeed.
\newblock \url{https://www.deepspeed.ai/}, 2023.

\bibitem{dhariwal2021diffusion}
Prafulla Dhariwal and Alexander Nichol.
\newblock Diffusion models beat gans on image synthesis.
\newblock {\em Advances in neural information processing systems}, 34:8780--8794, 2021.

\bibitem{dosovitskiy2020image}
Alexey Dosovitskiy, Lucas Beyer, Alexander Kolesnikov, Dirk Weissenborn, Xiaohua Zhai, Thomas Unterthiner, Mostafa Dehghani, Matthias Minderer, Georg Heigold, Sylvain Gelly, et~al.
\newblock An image is worth 16x16 words: Transformers for image recognition at scale.
\newblock {\em arXiv preprint arXiv:2010.11929}, 2020.

\bibitem{fang2023structural}
Gongfan Fang, Xinyin Ma, and Xinchao Wang.
\newblock Structural pruning for diffusion models.
\newblock {\em arXiv preprint arXiv:2305.10924}, 2023.

\bibitem{gagniuc2017markov}
Paul~A Gagniuc.
\newblock {\em Markov chains: from theory to implementation and experimentation}.
\newblock John Wiley \& Sons, 2017.

\bibitem{NIPS2014_5ca3e9b1}
Ian Goodfellow, Jean Pouget-Abadie, Mehdi Mirza, Bing Xu, David Warde-Farley, Sherjil Ozair, Aaron Courville, and Yoshua Bengio.
\newblock Generative adversarial nets.
\newblock In {\em Advances in Neural Information Processing Systems}, 2014.

\bibitem{grave2017improving}
Edouard Grave, Armand Joulin, and Nicolas Usunier.
\newblock Improving neural language models with a continuous cache.
\newblock In {\em International Conference on Learning Representations}, 2017.

\bibitem{gunasekaran2022cocktail}
Jashwant~Raj Gunasekaran, Cyan~Subhra Mishra, Prashanth Thinakaran, Bikash Sharma, Mahmut~Taylan Kandemir, and Chita~R Das.
\newblock Cocktail: A multidimensional optimization for model serving in cloud.
\newblock In {\em 19th USENIX Symposium on Networked Systems Design and Implementation (NSDI 22)}, pages 1041--1057, 2022.

\bibitem{hessel2021clipscore}
Jack Hessel, Ari Holtzman, Maxwell Forbes, Ronan Le~Bras, and Yejin Choi.
\newblock Clipscore: A reference-free evaluation metric for image captioning.
\newblock In {\em Proceedings of the 2021 Conference on Empirical Methods in Natural Language Processing}, pages 7514--7528, 2021.

\bibitem{heusel2017gans}
Martin Heusel, Hubert Ramsauer, Thomas Unterthiner, Bernhard Nessler, and Sepp Hochreiter.
\newblock Gans trained by a two time-scale update rule converge to a local nash equilibrium.
\newblock {\em Advances in neural information processing systems}, 30, 2017.

\bibitem{ho2020denoising}
Jonathan Ho, Ajay Jain, and Pieter Abbeel.
\newblock Denoising diffusion probabilistic models.
\newblock {\em Advances in neural information processing systems}, 33:6840--6851, 2020.

\bibitem{hu2021lora}
Edward~J Hu, Phillip Wallis, Zeyuan Allen-Zhu, Yuanzhi Li, Shean Wang, Lu~Wang, Weizhu Chen, et~al.
\newblock Lora: Low-rank adaptation of large language models.
\newblock In {\em International Conference on Learning Representations}, 2021.

\bibitem{huggingface}
Jina.ai.
\newblock Benchmark vector search databases with one million data.
\newblock \url{https://jina.ai/news/benchmark-vector-search-databases-with-one-million-data/}, 2022.

\bibitem{kingma2013auto}
Diederik~P Kingma and Max Welling.
\newblock Auto-encoding variational bayes.
\newblock {\em arXiv preprint arXiv:1312.6114}, 2013.

\bibitem{kirstain2023pick}
Yuval Kirstain, Adam Polyak, Uriel Singer, Shahbuland Matiana, Joe Penna, and Omer Levy.
\newblock Pick-a-pic: An open dataset of user preferences for text-to-image generation.
\newblock {\em arXiv preprint arXiv:2305.01569}, 2023.

\bibitem{kong2021fast}
Zhifeng Kong and Wei Ping.
\newblock On fast sampling of diffusion probabilistic models.
\newblock In {\em ICML Workshop on Invertible Neural Networks, Normalizing Flows, and Explicit Likelihood Models}, 2021.

\bibitem{kuhn-1988-speech}
Roland Kuhn.
\newblock Speech recognition and the frequency of recently used words: A modified {M}arkov model for natural language.
\newblock In {\em {C}oling {B}udapest 1988 Volume 1: {I}nternational {C}onference on {C}omputational {L}inguistics}, 1988.

\bibitem{kumar2019accelerating}
Adarsh Kumar, Arjun Balasubramanian, Shivaram Venkataraman, and Aditya Akella.
\newblock Accelerating deep learning inference via freezing.
\newblock In {\em 11th USENIX Workshop on Hot Topics in Cloud Computing (HotCloud 19)}, 2019.

\bibitem{lai2023modelkeeper}
Fan Lai, Yinwei Dai, Harsha~V Madhyastha, and Mosharaf Chowdhury.
\newblock $\{$ModelKeeper$\}$: Accelerating $\{$DNN$\}$ training via automated training warmup.
\newblock In {\em NSDI}, 2023.

\bibitem{lee2018pretzel}
Yunseong Lee, Alberto Scolari, Byung-Gon Chun, Marco~Domenico Santambrogio, Markus Weimer, and Matteo Interlandi.
\newblock $\{$PRETZEL$\}$: Opening the black box of machine learning prediction serving systems.
\newblock In {\em OSDI}, 2018.

\bibitem{li2023qdiffusion}
Xiuyu Li, Yijiang Liu, Long Lian, Huanrui Yang, Zhen Dong, Daniel Kang, Shanghang Zhang, and Kurt Keutzer.
\newblock Q-diffusion: Quantizing diffusion models.
\newblock {\em arXiv}, 2023.

\bibitem{liu2022pseudo}
Luping Liu, Yi~Ren, Zhijie Lin, and Zhou Zhao.
\newblock Pseudo numerical methods for diffusion models on manifolds.
\newblock {\em arXiv preprint arXiv:2202.09778}, 2022.

\bibitem{malkov2018efficient}
Yu~A Malkov and Dmitry~A Yashunin.
\newblock Efficient and robust approximate nearest neighbor search using hierarchical navigable small world graphs.
\newblock {\em IEEE transactions on pattern analysis and machine intelligence}, 42(4):824--836, 2018.

\bibitem{meng2023distillation}
Chenlin Meng, Robin Rombach, Ruiqi Gao, Diederik Kingma, Stefano Ermon, Jonathan Ho, and Tim Salimans.
\newblock On distillation of guided diffusion models.
\newblock In {\em CVPR}, 2023.

\bibitem{paszke2019pytorch}
Adam Paszke, Sam Gross, Francisco Massa, Adam Lerer, James Bradbury, Gregory Chanan, Trevor Killeen, Zeming Lin, Natalia Gimelshein, Luca Antiga, et~al.
\newblock Pytorch: An imperative style, high-performance deep learning library.
\newblock {\em Advances in neural information processing systems}, 32, 2019.

\bibitem{scikit-learn}
F.~Pedregosa, G.~Varoquaux, A.~Gramfort, V.~Michel, B.~Thirion, O.~Grisel, M.~Blondel, P.~Prettenhofer, R.~Weiss, V.~Dubourg, J.~Vanderplas, A.~Passos, D.~Cournapeau, M.~Brucher, M.~Perrot, and E.~Duchesnay.
\newblock Scikit-learn: Machine learning in {P}ython.
\newblock {\em Journal of Machine Learning Research}, 12:2825--2830, 2011.

\bibitem{podell2023sdxl}
Dustin Podell, Zion English, Kyle Lacey, Andreas Blattmann, Tim Dockhorn, Jonas M{\"u}ller, Joe Penna, and Robin Rombach.
\newblock Sdxl: improving latent diffusion models for high-resolution image synthesis.
\newblock {\em arXiv preprint arXiv:2307.01952}, 2023.

\bibitem{radford2021learning}
Alec Radford, Jong~Wook Kim, Chris Hallacy, Aditya Ramesh, Gabriel Goh, Sandhini Agarwal, Girish Sastry, Amanda Askell, Pamela Mishkin, Jack Clark, et~al.
\newblock Learning transferable visual models from natural language supervision.
\newblock In {\em International conference on machine learning}, pages 8748--8763. PMLR, 2021.

\bibitem{rombach2022high}
Robin Rombach, Andreas Blattmann, Dominik Lorenz, Patrick Esser, and Bj{\"o}rn Ommer.
\newblock High-resolution image synthesis with latent diffusion models.
\newblock In {\em Proceedings of the IEEE/CVF conference on computer vision and pattern recognition}, pages 10684--10695, 2022.

\bibitem{ronneberger2022convolutional}
O~Ronneberger, P~Fischer, and T~Brox.
\newblock Convolutional networks for biomedical image segmentation.
\newblock In {\em Medical Image Computing and Computer-Assisted Intervention--MICCAI 2015 Conference Proceedings}, 2022.

\bibitem{saharia2022photorealistic}
Chitwan Saharia, William Chan, Saurabh Saxena, Lala Li, Jay Whang, Emily~L Denton, Kamyar Ghasemipour, Raphael Gontijo~Lopes, Burcu Karagol~Ayan, Tim Salimans, et~al.
\newblock Photorealistic text-to-image diffusion models with deep language understanding.
\newblock {\em Advances in Neural Information Processing Systems}, 35:36479--36494, 2022.

\bibitem{salimans2021progressive}
Tim Salimans and Jonathan Ho.
\newblock Progressive distillation for fast sampling of diffusion models.
\newblock In {\em International Conference on Learning Representations}, 2021.

\bibitem{schubert2017dbscan}
Erich Schubert, J{\"o}rg Sander, Martin Ester, Hans~Peter Kriegel, and Xiaowei Xu.
\newblock Dbscan revisited, revisited: why and how you should (still) use dbscan.
\newblock {\em ACM Transactions on Database Systems (TODS)}, 42(3):1--21, 2017.

\bibitem{shih2023parallel}
Andy Shih, Suneel Belkhale, Stefano Ermon, Dorsa Sadigh, and Nima Anari.
\newblock Parallel sampling of diffusion models, 2023.

\bibitem{song2020denoising}
Jiaming Song, Chenlin Meng, and Stefano Ermon.
\newblock Denoising diffusion implicit models.
\newblock In {\em International Conference on Learning Representations}, 2020.

\bibitem{wang2023exploring}
Jianyi Wang, Kelvin~CK Chan, and Chen~Change Loy.
\newblock Exploring clip for assessing the look and feel of images.
\newblock In {\em Proceedings of the AAAI Conference on Artificial Intelligence}, volume~37, pages 2555--2563, 2023.

\bibitem{wang2023tabi}
Yiding Wang, Kai Chen, Haisheng Tan, and Kun Guo.
\newblock Tabi: An efficient multi-level inference system for large language models.
\newblock In {\em Proceedings of the Eighteenth European Conference on Computer Systems}, pages 233--248, 2023.

\bibitem{wang2022diffusiondb}
Zijie~J Wang, Evan Montoya, David Munechika, Haoyang Yang, Benjamin Hoover, and Duen~Horng Chau.
\newblock Diffusiondb: A large-scale prompt gallery dataset for text-to-image generative models.
\newblock {\em arXiv preprint arXiv:2210.14896}, 2022.

\bibitem{wu2023fast}
Zike Wu, Pan Zhou, Kenji Kawaguchi, and Hanwang Zhang.
\newblock Fast diffusion model, 2023.

\bibitem{yadwadkar2019case}
Neeraja~J Yadwadkar, Francisco Romero, Qian Li, and Christos Kozyrakis.
\newblock A case for managed and model-less inference serving.
\newblock In {\em Proceedings of the Workshop on Hot Topics in Operating Systems}, pages 184--191, 2019.

\bibitem{yang2021segcache}
Juncheng Yang, Yao Yue, and Rashmi Vinayak.
\newblock Segcache: a memory-efficient and scalable in-memory key-value cache for small objects.
\newblock In {\em 18th USENIX Symposium on Networked Systems Design and Implementation (NSDI 21)}, pages 503--518, 2021.

\bibitem{yin2021towards}
Miao Yin, Yang Sui, Siyu Liao, and Bo~Yuan.
\newblock Towards efficient tensor decomposition-based dnn model compression with optimization framework.
\newblock In {\em CVPR}, 2021.

\bibitem{zhang2019mark}
Chengliang Zhang, Minchen Yu, Wei Wang, and Feng Yan.
\newblock $\{$MArk$\}$: Exploiting cloud services for $\{$Cost-Effective$\}$,$\{$SLO-Aware$\}$ machine learning inference serving.
\newblock In {\em 2019 USENIX Annual Technical Conference (USENIX ATC 19)}, pages 1049--1062, 2019.

\bibitem{zhang2023text}
Chenshuang Zhang, Chaoning Zhang, Mengchun Zhang, and In~So Kweon.
\newblock Text-to-image diffusion model in generative ai: A survey.
\newblock {\em arXiv preprint arXiv:2303.07909}, 2023.

\bibitem{zhang2023prospect}
Yuxin Zhang, Weiming Dong, Fan Tang, Nisha Huang, Haibin Huang, Chongyang Ma, Tong-Yee Lee, Oliver Deussen, and Changsheng Xu.
\newblock Prospect: Expanded conditioning for the personalization of attribute-aware image generation.
\newblock {\em arXiv preprint arXiv:2305.16225}, 2023.

\bibitem{zheng2023fast}
Hongkai Zheng, Weili Nie, Arash Vahdat, Kamyar Azizzadenesheli, and Anima Anandkumar.
\newblock Fast sampling of diffusion models via operator learning.
\newblock In {\em International Conference on Machine Learning}, pages 42390--42402. PMLR, 2023.

\end{thebibliography}
